\documentclass[11pt, a4paper, logo, onecolumn]{preprint}

\usepackage[all]{hypcap}
\usepackage{xspace}
\usepackage[capitalize,noabbrev]{cleveref}
\usepackage{subcaption}
\usepackage{wrapfig}
\usepackage{lipsum}
\usepackage{multirow}
\usepackage{tabularx}
\usepackage{algorithm}
\usepackage{algpseudocode}

\usepackage{url}
\usepackage{float}
\usepackage{tikz}
\usepackage{tabularx}
\usepackage{array}
\newcolumntype{Y}{>{\centering\arraybackslash}X}
\usepackage{multirow} 
\usepackage{makecell}
\usepackage{subcaption}
\usepackage{footnote}
\usepackage{hyperref}

\usepackage{amsmath}

\makeatletter
\def\mathcolor#1#{\@mathcolor{#1}}
\def\@mathcolor#1#2#3{%
  \protect\leavevmode
  \begingroup
    \color#1{#2}#3%
  \endgroup
}
\makeatother

\title{AnyDexGrasp: General Dexterous Grasping for Different Hands with Human-level Learning Efficiency}
\runningtitle{AnyDexGrasp: General Dexterous Grasping for Different Hands with Human-level Learning Efficiency}

\reportnumber{} 

\author[1]{Hao-Shu Fang}
\author[1]{Hengxu Yan}
\author[1]{Zhenyu Tang}
\author[1]{Hongjie Fang}
\author[1]{Chenxi Wang}
\author[2]{Cewu Lu}

\affil[1]{Department of Computer Science, Shanghai Jiao Tong University}
\affil[2]{School of Artificial Intelligence, Shanghai Jiao Tong University}

\email{ \href{mailto:fhaoshu@gmail.com}{fhaoshu@gmail.com}, \{hengxuyan, tang\_zhenyu, galaxies, wcx1997, lucewu\}@sjtu.edu.cn}

\begin{abstract}
We introduce an efficient approach for learning dexterous grasping with minimal data, advancing robotic manipulation capabilities across different robotic hands. Unlike traditional methods that require millions of grasp labels for each robotic hand, our method achieves high performance with human-level learning efficiency: only hundreds of grasp attempts on 40 training objects. The approach separates the grasping process into two stages: first, a universal model maps scene geometry to intermediate contact-centric grasp representations, independent of specific robotic hands. Next, a unique grasp decision model is trained for each robotic hand through real-world trial and error, translating these representations into final grasp poses. Our results show a grasp success rate of 75-95\% across three different robotic hands in real-world cluttered environments with over 150 novel objects, improving to 80-98\% with increased training objects. This adaptable method demonstrates promising applications for humanoid robots, prosthetics, and other domains requiring robust, versatile robotic manipulation. Project website: \href{https://graspnet.net/anydexgrasp/}{https://graspnet.net/anydexgrasp/}.
\end{abstract}

\begin{document}

\maketitle

\section*{Summary}
A general framework for learning visually guided dexterous grasping across various robotic hands, achieving human-level learning efficiency and robust performance in cluttered environments.

\section{Introduction}
Grasping, as a fundamental problem of prehensile manipulation, holds significant importance in robotics. Over the past decades, diverse mechanical structures for robotic hands have been developed. Visually guided dexterous grasping is in high demand to enable robots to interact effectively with their environments. This ability also plays a crucial role in the context of intelligence. Throughout human evolution, early humans developed the capability for precise grip~\cite{doi:10.1126/science.1261735}, which enabled tool use and is believed to have facilitated the evolution of the human species~\cite{almecija2010early, kivell2015evidence}. From the perspectives of both advancing robotics and promoting embodied intelligence, it is essential to design a learning framework that efficiently equips different robotic hands with visually guided dexterous grasping capabilities.

To make such a grasping system practically useful, it should use a single commodity camera, observe environments with cluttered objects, handle perception noise and generate a set of dexterous grasp poses that can be selected by subsequent tasks. Due to the challenges of the problem, early research focused on generating dexterous grasp poses given a single, complete object mesh, utilizing either analytical~\cite{miller2004graspit, rosales2011synthesizing, liu2021synthesizing, liu2020deep} or learning-based approaches~\cite{li2023gendexgrasp}. The idea is to decouple the grasping system into 6D pose estimation and grasp poses generation based on the object CAD model. However, the requirement for the object mesh limits its ability to handle new object shapes.

It is challenging to detect grasp poses for unseen objects based on partial-view perception. Some recent methodologies pursue mesh completion using partial point clouds~\cite{lundell2021ddgc, wei2022dvgg, wei2024learning}, followed by grasps generation on the complete mesh. However, the error introduced by perception noise and mesh completion often results in inaccurate grasp analysis.  An increasing amount of research has attempted to learn the mapping from raw partial observation to grasp poses within a single network. Due to the highly nonlinear property of this mapping, extensive training data is required. Two data sources are commonly adopted: human grasping demonstrations~\cite{gupta2016learning, christen2019guided, qin2022dexmv, mandikal2022dexvip, wei2024learning, shaw2024learning} or data from simulated environments\cite{brahmbhatt2019contactgrasp, corona2020ganhand, grady2021contactopt,li2023gendexgrasp, wang2023dexgraspnet, lum2024dextrahg,
singh2025dextrahrgb}. However, both methods have their limitations. The former approach struggles to accurately capture human hand gestures and is confined to robotic hands resembling human anatomy. The latter requires substantial effort to build the simulation environment, annotate grasp poses by analytical models or trial and error, and transfer algorithms from simulation to the real world. These challenges limit current grasping systems to simple scenarios, typically involving a single object at a time from a limited set.  No prior work demonstrates robust grasping in cluttered environments from partial-view perception in the real world.

Most critically, even if these challenges are overcome, the policy obtained with substantial efforts is only suitable for a specific robotic hand each time. The  end-to-end learning paradigm implicitly encodes the information about hand kinematic structure, relevant state information and grasp quality in the weights, making it difficult for models to share computation between different hands. Consequently, we need to repeat the tedious data generation and policy training pipeline for each hand.

We identified two main bottlenecks for efficiently learning visually guided dexterous grasping for different robotic hands: the requirements for extensive training data for each hand, and the inability to share  computation across different hands. These bottlenecks arise from attempting to learn the mapping from raw observation to grasp poses with an end-to-end network. In this paper, we revisit this paradigm. We hypothesize that if there exists a low-dimensional intermediate state space that encapsulates grasp information, then the mapping from this state space to grasp poses can be learned more efficiently than the original mapping, requiring less training data. Moreover, if such a state space is transferable across different robotic hands, it could be shared without the need to retrain a state estimator each time. Note that such a state space should not require object knowledge during inference, in order to generalize to unseen objects.

Recognizing this potential, we aimed to identify such a state space. For the grasping problem, the robot needs to decide its grasp forces on each finger, based on the grasp matrix, surface normals of contact points and friction coefficient~\cite{dai2018synthesis}. From visual perception, the information we can extract is the positions and normals of potential contact points, where positions are linked to the grasp matrix and normals determine the orientations of friction cones. Based on this observation, we introduce a novel intermediate representation for multi-finger grasping, referred to as the Contact-centric Grasp Representation (CGR), which encapsulates contact information on the object's surface and possesses SE(3)-equivalent property.

Based on this representation, we present AnyDexGrasp, a novel methodology that can effectively learn dexterous grasping for different hands on a modest set of training objects. In this method, the multi-finger grasp detection problem that maps raw perception to grasp poses is divided into two steps. In the first step, we train a general representation model that maps single-view partial observations to contact-centric grasp representations. A large-scale dataset is annotated to train this model. After training, it can be applied to different hands without fine-tuning. In the second step, we map the contact-centric grasp representations to a set of grasp proposals through a hand-specific mapping, and then learn a hand-specific classifier to evaluate each grasp proposal. This classifier takes a contact-centric grasp representation and a grasp proposal as input and maps them to the probability of grasp success. The training data is collected by real-world trial and error.  We empirically observed that this mapping is significantly easier to learn, requiring merely hundreds of trial-and-error attempts. It dramatically reduces the cost of real-world learning and allows our approach to work for different types of robotic hands efficiently.

We evaluate the effectiveness of our method using three different robotic hands, each featuring three to five fingers. Our system is first trained on 144 objects, with approximately 2,000 to 8,000 grasp attempts, depending on the robotic hand. On a diverse set of 150 previously unseen objects, including deformable and adversarial items, our approach achieves an average grasp success rate ranging from 80\% to 98\% across different hands.  Notably, this performance is achieved in cluttered scenarios, demonstrating the effectiveness of our approach.

In addition we explore further reductions in training samples required for our grasp learning paradigm. We limit the training objects to 40 and reduce the grasp attempts to approximately 400 to 1,000 depending on the robotic hand. Even with this limited amount of training data, our system consistently achieves grasp success rates ranging from 75\% to 95\% during real-world testing. Notably, our experiments also highlight the potential for further reductions of training samples, with the ability to decrease the training object number to 30 and the total grasp attempts to 200 for a three-finger hand, without decreasing the grasp performance by a large margin.  Such learning efficiency allows robots to master visually guided grasping in a matter of hours in the real world, surpassing the learning efficiency of human infants.

We conduct a series of analyses to clarify why our two step learning method is so efficient. In the first step, we perform a geometry coverage analysis, showcasing that by scaling up data in the correct dimension, the local geometries on just 40 objects can effectively cover a wide range of unseen objects. This explains the generalization capabilities with a small number of training objects. In the second step, we provide various perspectives illustrating how our proposed contact-centric grasp representation serves as a robust state space for grasp decision, which allows the model to learn from just hundreds of real-world trial-and-error attempts.

This paper represents a significant step toward the efficient realization of dexterous robotic grasping, with the potential to revolutionize various applications, from advanced humanoid robots to prosthetic hands.

\begin{figure}[htbp]
    \centering
    \includegraphics[width=1\textwidth]{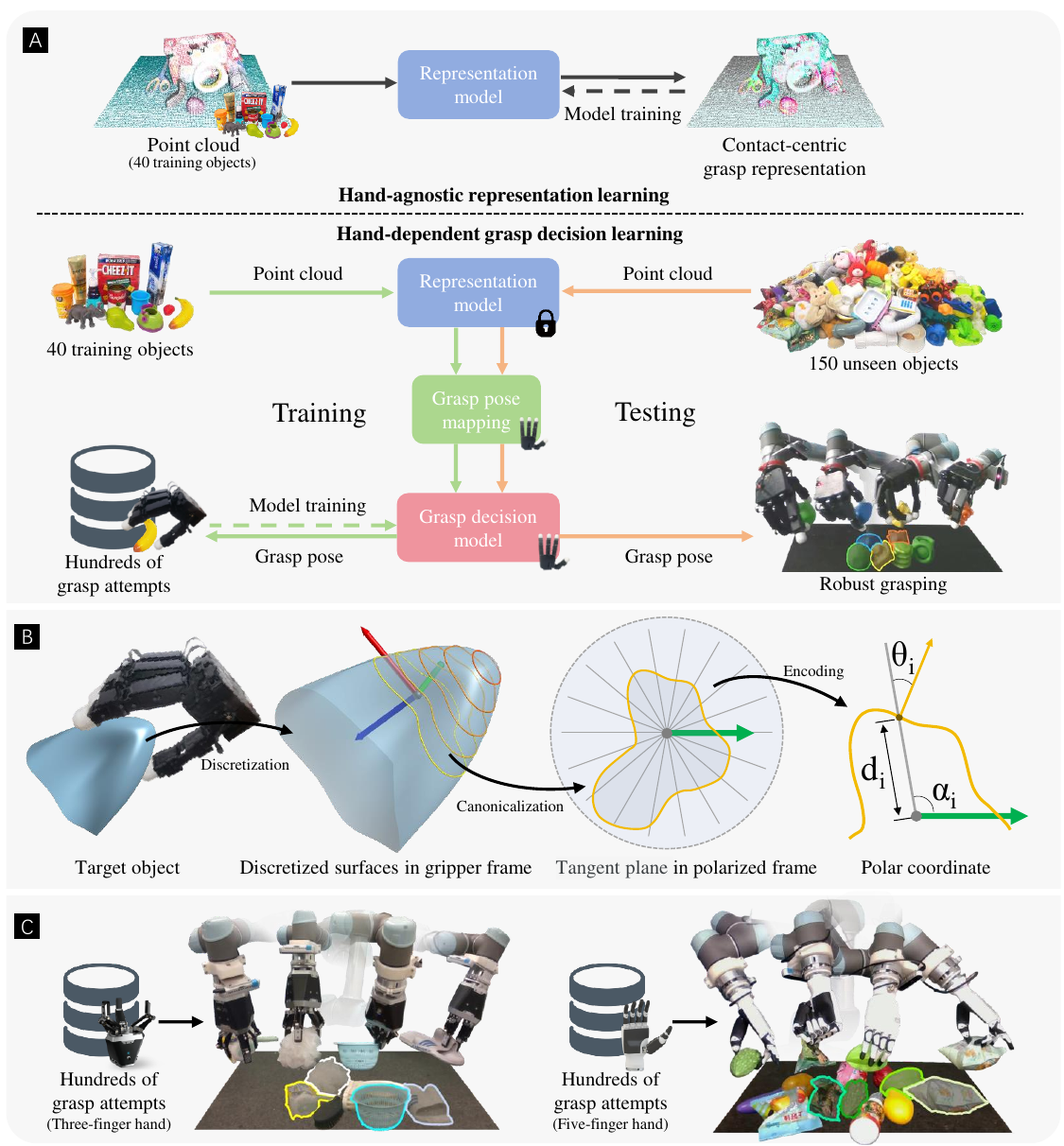}
    \caption{\textbf{The overview of our method.} \textbf{(A)}: Our method consists of two steps. The first step is to train a representation model on partial-view point cloud. The training set only consists of 40 objects. The second step would fix the representation model, and train a grasp decision model that takes the grasp-centric contact representation as input and outputs the grasp success score, based on hundreds of real-world trial-and-error attempts. The grasp algorithm is tested thoroughly on hundreds of unseen objects. \textbf{(B)}: Illustration of contact-centric grasp representation. A local geometry is discrete into several tangent planes along the approach direction of a robotic hand. Each tangent surface is transformed into the polarized coordinate frame of the robotic hand. The shape of the surface is encoded into discretized points and normal representation in the polar coordinate. \textbf{(C)}: Our experiments are also carried out on a three-finger hand and a five-finger hand and demonstrate excellent performance.}
    \label{fig:teaser}
\end{figure}

\section{Preliminary}
\subsection{Formulation}

A multi-finger grasp pose $g$ is formally defined as:
\begin{equation}\label{multi-finger definition1}
    g = [\mathbf{R}\ \mathbf{t}\ \mathbf{q}],
\end{equation}
where $\mathbf{R} \in \mathbb{R}^{3\times3}$ represents the robotic hand's rotation, $\mathbf{t} \in \mathbb{R}^{3\times1}$ denotes the hand's translation, and $\mathbf{q} \in \mathbb{R}^{n\times1}$ characterizes the joint configuration of a $n$-DoF multi-finger hand. The goal of the grasp pose detection problem is to predict a set of grasp poses from a scene perception. Conventionally, data-driven methods have employed a single network $f(\cdot)$ to map the partial point cloud of the scene $\mathcal{P} \in \mathbb{R}^{k\times3}$ to a set of candidate poses, $\mathbf{G}=\left\{g_i\right\}_{i=1}^{|\mathbf{G}|}$. In contrast, our approach decouples the mapping into two distinct steps: a state embedding step and a grasp decision step.

In the state embedding step, we extract a collection of contact-centric grasp representations from the partial point cloud $\mathcal{P}$. This is achieved by using a hand-agnostic representation model $\Phi(\cdot)$, which generates the scene representation $\mathcal{R}$:
\begin{equation}
\mathcal{R} = \Phi(\mathcal{P}),
\end{equation}
where $\mathcal{R} = \left\{r_j\right\}_{j=1}^{|\mathcal{R}|}$ is a set of contact-centric grasp representations.

The grasp decision step consists of two distinct procedures: a mapping process that converts contact-centric grasp representations into a set of candidate grasp poses (referred to as grasp candidates) and a quality estimation process for each candidate. For each grasp representation $r_j$, we generate a set of grasp candidates based on the specific robotic hand. A hand-dependent mapping function $\mathcal{K}(\cdot)$ takes a grasp representation $r_j$ and a hand specification $h$ as input, and output $\mathbf{G}_j$:
\begin{equation}
\mathbf{G}_j = \mathcal{K}(r_j, h),
\end{equation}
where $\mathbf{G}_j = \left\{g^{(i)}_j\right\}_{i=1}^{|\mathbf{G}_j|}$ denotes the set of grasp candidates for each $r_j$.

To estimate the quality of a grasp, we use a hand-dependent grasp decision model $\Psi(\cdot)$, which predicts the probability of success $\beta$ given a grasp representation $r_j$, a grasp pose $g^{(i)}_j$, and a hand specification $h$:
\begin{equation}
\beta = \Psi(r_j, g^{(i)}_j, h).
\end{equation}

\textbf{Objective:} Our goal is to find a set of grasp poses $\mathbf{G}^*$ that maximizes the grasp success rate given a desired number of grasp poses $K$:

\begin{align}
\mathbf{G}^* & =  \argmax_{\mathbf{G} \subset \bigcup_j \mathbf{G}_j, \; |\mathbf G|=K}\mathbb{E}_{r_j\in \mathcal{R},\; g_j^{(i)}\in \mathcal{K}(r_j, h) \cap \mathbf{G}}[\Psi(r_j, g_j^{(i)}, h)].
\end{align}

Figure~\ref{fig:teaser}A shows the pipeline of our methodology.

\subsection{Contact-centric Grasp Representation} \label{antipodal_representaiton}
We initiate our approach with the development of a contact-centric grasp representation. Initially, consider a 2D object, we can represent it as a set comprising surface points and their corresponding normals:
\begin{equation}
r_{2d} = \{(p_i,n_i) \mid i=1,2,\cdots,N\}.
\end{equation}
In this representation, $p_i$ denotes the position of a surface point, and $n_i$ represents the normal vector associated with that surface point. For clarity, the object's surface is discretized into $N$ bins.
 
For the task of grasp pose detection, it is common to represent the object shape in a local coordinate frame~\cite{gpd,mousavian2019graspnet}, as the classification of grasp quality depends primarily on the geometry within a localized area. This step, referred to as canonicalization, equips the representation with SE(3)-equivalent property and makes subsequent learning easier. For the 2D example, when we employ a polar coordinate system and sample the pole coordinate $\mathbf{t}_{2d}$ and the polar axis $\mathbf{R}_{2d}$, the discrete object shape representation is refactored accordingly. In this system, a surface point $p_i$ is represented by an angle $\alpha_i$ from the polar axis and a distance $d_i$ from the pole. Additionally, the surface normal is encoded as the angle between the normal $n_i$ and $\alpha_i$:
\begin{align}
r_{2d} = \bigg\{ (\alpha_i, d_i, \theta_i) \, \mid \, & i = 1, 2, \ldots, N, \, p'_i = \mathbf{R}_{2d}(p_i - \mathbf{t}_{2d}), \, n'_i = \mathbf{R}_{2d}n_i, \nonumber \\
& \alpha_{i} = \frac{p'_{i}}{\|p'_{i}\|},  d_{i} = \|p'_{i}\|, \nonumber \\
& \theta_{i} = \arccos\left(\frac{\alpha_{i} \cdot n'_{i}}{\|\alpha_{i}\|\|n'_{i}\|}\right) ; \mathbf{R}_{2d},  \mathbf{t}_{2d} \bigg\}.
\end{align}

A benefit of adopting a polar coordinate system is that the in-plane rotation angles $\{\alpha_{i}\}$ can be uniformly sampled across the polar angle range, resulting in constant values for $\{\alpha_{i}\}$ across different representations. Therefore, we move $\{\alpha_{i}\}$ to the right side of the set notation to make the representation more compact. Since the values of $d_i$ and $\theta_i$ depend on $\alpha_{i}$, we rewrite them as $d_{\alpha_{i}}$ and $\theta_{\alpha_{i}}$:
\begin{align} 
r_{2d} = \bigg\{ (d_{\alpha_i}, \theta_{\alpha_i}) \mid \alpha_i = 0, \frac{2\pi}{N}, \ldots, \frac{2\pi (N-1)}{N} ; \mathbf{R}_{2d},  \mathbf{t}_{2d} \bigg\}. 
\end{align}

Extending this representation to a real-world 3D object and a 3D coordinate system with rotation $ \mathbf{R}_{3d}$ and translation $\mathbf{t}_{3d}$ involves decoupling the object's geometry along a chosen axis and composing multiple 2D representations.  By selecting a specific axis in the 3D coordinate system (\textit{e.g.}, the $z$-axis), we discretize the object along this axis into $M$ sections. Each section corresponds to a cross-sectional slice of the object at a particular coordinate along the axis.

Within each cross-sectional slice, the same polar coordinate system is employed as in the 2D case. We apply the same angular sampling and the local geometry is represented in terms of distance and normal angle at each sampled angle $\alpha_i$. The 3D representation is then formulated as:
\begin{align}
r_{3d} = \bigg\{ & (d_{\alpha_i}, \theta_{\alpha_i})_j \, \bigg| \, \alpha_i = 0, \frac{2\pi}{N}, \ldots, \frac{2\pi (N-1)}{N}, \, j = 1, 2, \ldots, M; \mathbf{R}_{3d},\mathbf{t}_{3d} \bigg\}
\label{eqn_r3d}
\end{align} 

In the following sections, we use $r$ as a shorthand for $r_{3d}$. In Figure~\ref{fig:teaser}B, we illustrate the process of representing a 3D geometry in the contact-centric representation format within a robotic hand's local coordinate frame.

\subsection{Robotic Hands and Grasp Types}

\begin{figure}[t]
    \centering
    \includegraphics[width=1\textwidth]{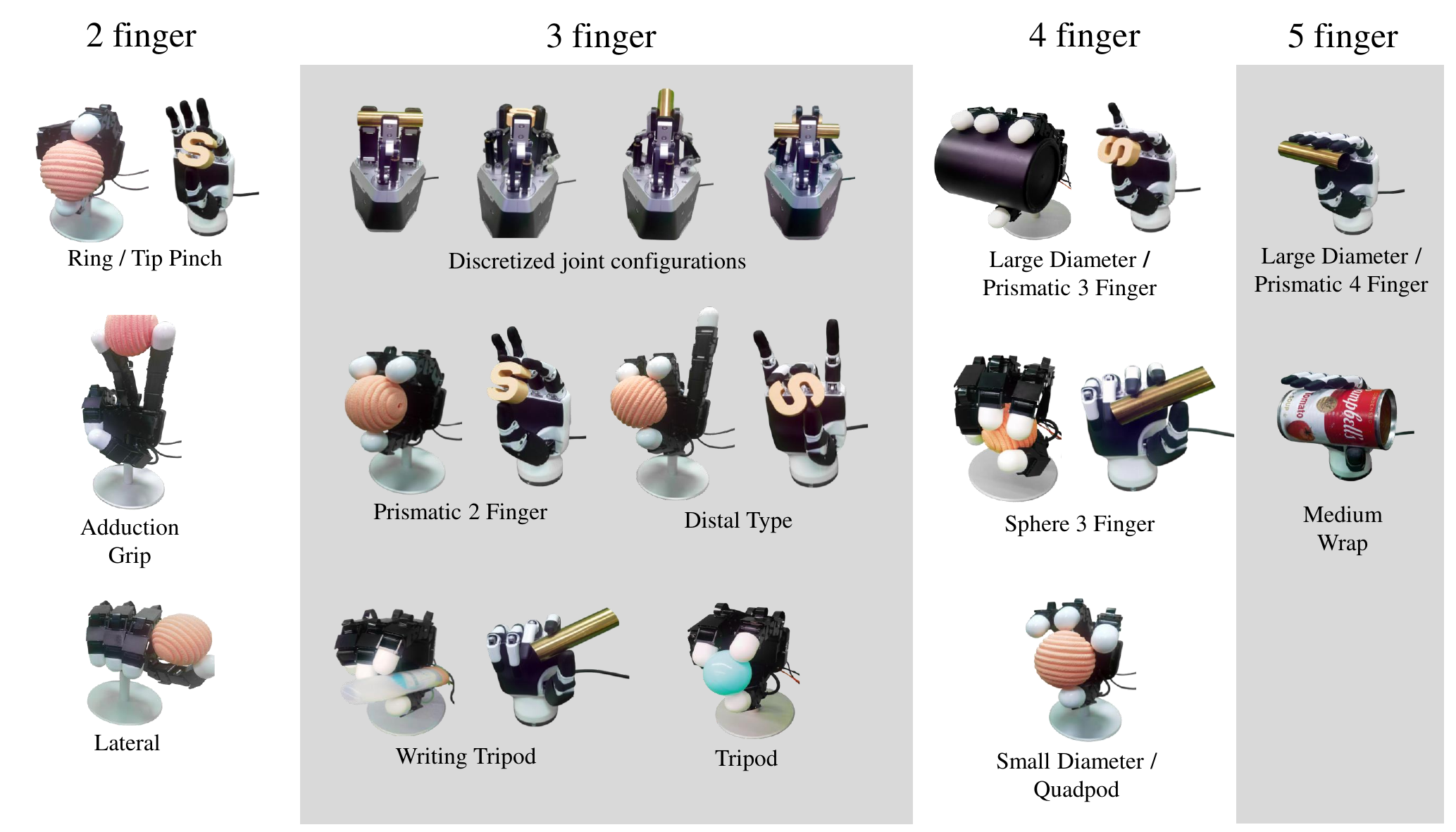}
    \caption{\textbf{Illustration of the predefined grasp types for three robotic hands.} The types are categorized by the number of fingers involved in the grasping procedure. Some types can be categorized into multiple taxonomies defined in previous literature~\cite{feix2015grasp} when the grasping depths differ.}
    \label{fig:hand_pose}
\end{figure}

In our experiments, we utilize three distinct robotic hands:
\begin{itemize}
    \item DH-3: A three-finger robotic hand comprises 4 degrees of freedom and 2 motors, operating in an underactuated manner.
    \item Allegro: A four-finger robotic hand comprises 16 degrees of freedom and 16 motors, designed for full actuation.
    \item Inspire: A five-finger robotic hand equipped with 12 degrees of freedom and 6 motors, operating in an underactuated manner.
\end{itemize}
These robotic hands represent a variety of applications, including industrial tasks, dexterous manipulation, and underactuated prosthetic hand functionalities.

One challenge with dexterous grasping is the complexity introduced by the high degrees of freedom in these robotic hands, which creates a vast joint configuration space. However, when humans grasp objects, we typically rely on only a small subset of these configurations, which can be categorized into specific taxonomies~\cite{cutkosky1989grasp,feix2015grasp}. To address this complexity and make grasp pose detection more manageable, we discretize the continuous joint configurations of the multi-fingered hands into a finite set of predefined grasp types. This is represented as $\mathbf{q} \in \{\mathbf{q}_1, ..., \mathbf{q}_c\}$, where $c$ denotes the total number of grasp types specific to each hand.

For the three-finger hand, we discretize the entire joint space into several bins, while for the four- and five-finger hands, we select grasp types from the human grasp taxonomy that can be executed by these dexterous robotic hands. The predefined grasp types are illustrated in Figure~\ref{fig:hand_pose}. While this approach simplifies the grasp pose detection process, it still provides sufficient flexibility for subsequent manipulation tasks.

It is important to note that these grasp types serve as anchor poses prior to contact. Once the hand reaches its target position, it undergoes a closure process, where the fingers progressively move toward each other until the forces exerted on the finger joints reach predefined limits.

\begin{figure}[t]
    \centering
    \includegraphics[width=1\textwidth]{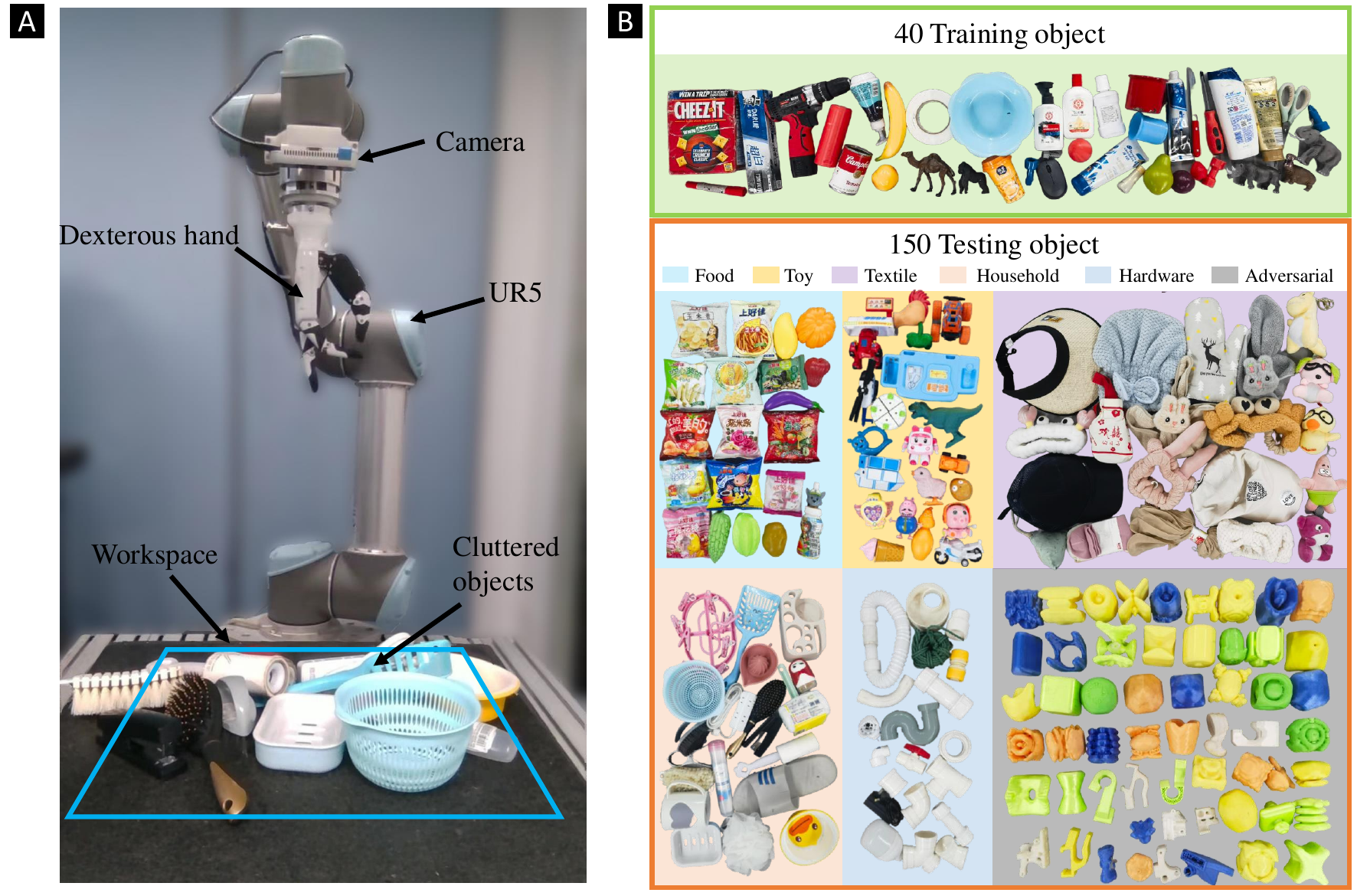}
    \caption{\textbf{Experimental setup.} \textbf{(A)}: Platform setting of our dexterous grasping experiments. \textbf{(B)}: Illustration of our 40 training objects and 150 testing objects. The testing objects are much more diverse than the training objects, including deformable and adversarial objects not presented in the training set.}
    \label{fig:objects}
\end{figure}

\subsection{Overview of Experiments}
To assess the performance of our multi-finger grasping model, we established a real-world experimental platform. Our hardware setup includes a UR5 robotic arm and an Intel RealSense D415 camera, positioned at the robot's end-effector. The initial camera pose is vertical to the table and is approximately 60 cm above it. Figure~\ref{fig:objects}A illustrates the setup of our robotic platform.

We first learn a general hand-agnostic representation model based on an offline annotated, large-scale dataset. Once the representation model is learned, we use the predicted contact-centric grasp representation as a new state space for the problem of multi-finger grasping. For each robotic hand, we can learn grasping in the real world directly through trial and error. We start with thousands of trial-and-error grasp attempts, and gradually reduce the number to hundreds of attempts in later experiments to demonstrate the efficiency of our learning paradigm. We also vary the number of training objects from 144 to 40 and even 30, to verify the generalization ability of our grasp system. 

To assess the multi-finger grasp performance thoroughly, we construct a comprehensive real-world test set, featuring objects commonly encountered in everyday life. These objects encompass diverse shapes, materials, and textures and are categorized into hardware, food, textile, household, toy, and adversarial items. The test set comprises nearly 150 objects ranging in size from $2.5 \times 2.5 \times 2.5$ cm$^3$ to $8 \times 8 \times 5$ cm$^3$.

During real-world testing, objects from each category are randomly placed on a table in a cluttered way, and the robots attempt to grasp all the objects and clear the table. This process is repeated twice for accuracy. We also establish a baseline that aligns the principal closing axis of the grasp types with antipodal grasp poses, followed by collision detection, to compare with our proposed method. The success rate is determined by dividing the number of successful grasp attempts by the total number of grasp attempts.

Ultimately, our grasp system is successfully evaluated on three different robotic hands, where the whole system is trained on a limited dataset comprising merely 40 objects and hundreds of grasp attempts, and tested on a broader spectrum of 150 previously unseen objects. Notably, it represents a pioneering achievement in the literature where a grasping algorithm is evaluated on a significantly larger set of objects than those included in its training dataset. The final training and testing objects are illustrated in Figure~\ref{fig:objects}B for reference.

\section{Results}
\subsection{Learning Dexterous Grasping in the Real World}
Based on the contact-centric grasp representations output by a trained representation model, we first employ a training set of 144 objects to train the grasp decision model. Approximately 1,000 grasp samples are collected for each grasp type of the DH-3 and Inspire Hand, and 200 grasp samples are collected for each grasp type of the Allegro Hand, forming the basis for our learning process. The amount of training objects and grasp samples would be gradually reduced in later sections to verify the effectiveness of our method.

\begin{figure}[t]
    \centering
    \includegraphics[width=\textwidth]{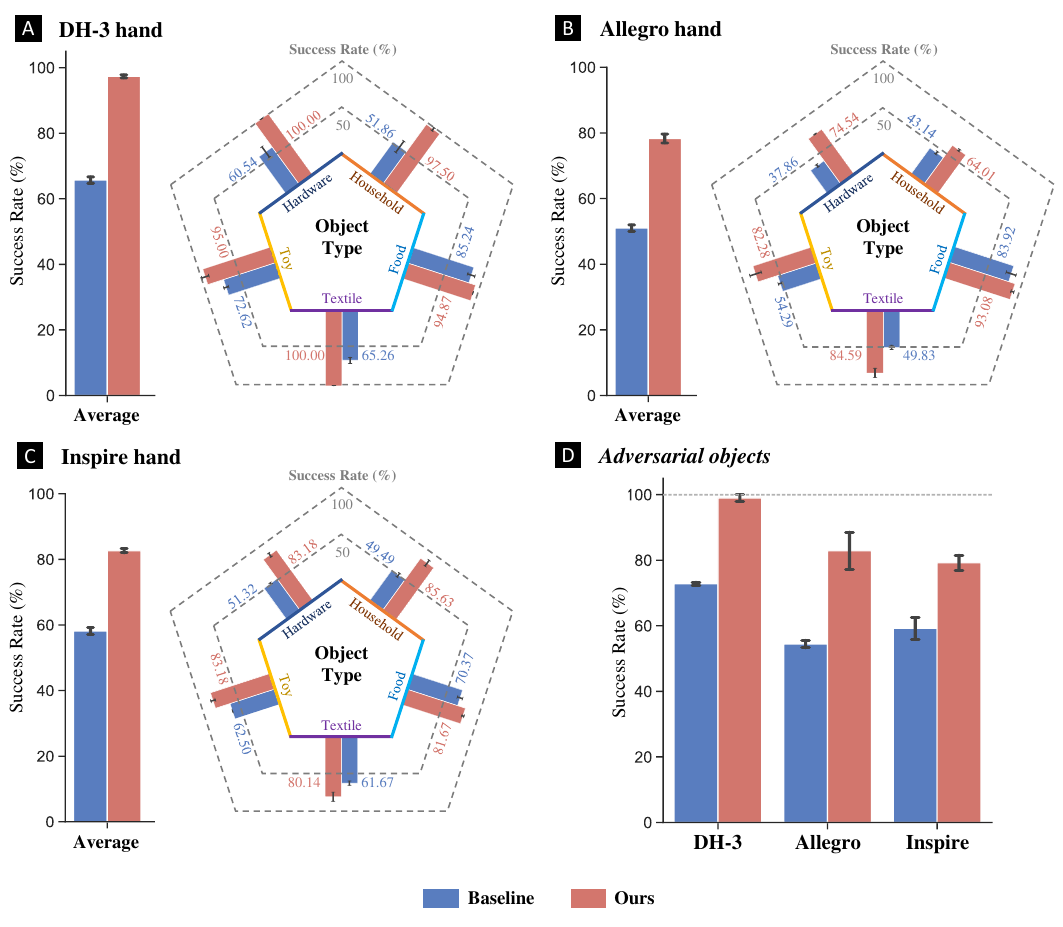}
    \caption{\textbf{Success rates on the testing set after training on abundant real-world data.} \textbf{(A)}: The averaged and detailed success rates of the DH-3 hand on five object categories commonly encountered in our daily activities. \textbf{(B)}: The averaged and detailed success rates of the Allegro hand. \textbf{(C)}: The averaged and detailed success rates of the Inspire hand. \textbf{(D)}: The success rates on the adversarial objects of three robotic hands.}
    \label{fig:main}
\end{figure}

\subsubsection*{Dexterous Grasping on Daily Objects}

We systematically evaluate the success rates of our approach on testing objects from the first five categories commonly encountered in our daily activities. The average success rates achieved by the three distinct robotic hands are 97\%, 78\%, and 83\%, respectively. Movies \hyperlink{movie_s1}{S1}, \hyperlink{movie_s2}{S2} and \hyperlink{movie_s3}{S3} record the grasping process. In contrast, the success rates of the baseline method using heuristic sampling and collision detection reach only 66\%, 51\%, and 58\%. A detailed breakdown of success rates for each object category is presented in Figure~\ref{fig:main}A. Compared to the baseline method, the substantial improvements across this extensive test set demonstrate the effectiveness of our proposed representation and approach.

Several noteworthy points are hereby highlighted. Firstly, the 3-finger gripper attains an average success rate of 97\% across over 100 real-world objects, surpassing even the performance of previous state-of-the-art parallel-gripper algorithm~\cite{fang2023anygrasp}. Secondly, for deformable objects within the textile and food categories, the grasp success rates across different grippers show no significant degradation. In some cases, they even slightly outperform the overall success rate, despite the absence of explicit training on deformable objects. This observation emphasizes the remarkable generalization capacity of data-driven methods. We observe that deformable objects tend to comply with the gripper during the grasping process, making them easier to be successfully grasped.

Regarding grasping speed, our system takes an average of 0.5 second to generate 200 grasp poses in a cluttered scene. Additionally, an extra collision detection step utilizing scene partial point cloud and hand mesh is performed. It takes 20 seconds on our CPU using Open3D library~\cite{Zhou2018}. Although this step could be accelerated through advanced collision detection technology or hardware acceleration, this aspect falls outside the scope of this paper.

\subsubsection*{Dexterous Grasping on Adversarial Objects}

In addition to daily objects, we extend our method’s evaluation to more challenging adversarial objects. These objects encompass 13 human-selected items from DexNet~\cite{dexnet2} and 49 program-generated objects from EGAD! evaluation set~\cite{morrison2020egad}, characterized by distinct shapes and varying grasp difficulties. Prior literature shows a performance degradation of parallel grasping on adversarial versus daily objects~\cite{fang2023anygrasp}. To the best of our knowledge, this is the first comprehensive evaluation of a multi-finger grasping algorithm on adversarial objects in real-world scenarios.

Success rates for the three distinct robotic hands are reported in Figure~\ref{fig:main}B, where our system achieves 99\%, 82\%, and 79\% success rates, respectively. Movies \hyperlink{movie_s4}{S4}, \hyperlink{movie_s5}{S5} and \hyperlink{movie_s6}{S6} record the grasping process.  In contrast, the baseline method achieves success rates of 72\%, 54\%, and 59\%. Remarkably, the performance on adversarial objects is on par with daily objects for all the robotic hands, highlighting the promising generalization ability of our dexterous grasping system. It surprises us since previous results from parallel grippers show a dramatic performance degradation. We presume that the additional fingers can improve the grasping ability, and the adversarial objects designed for parallel grippers do not pose significant challenges in multi-finger cases. 

In the subsequent sections, unless otherwise stated, we proceed to conduct experiments on all 150 objects including daily ones and adversarial ones.

\begin{figure}[t]
    \centering
    \includegraphics[width=\textwidth]{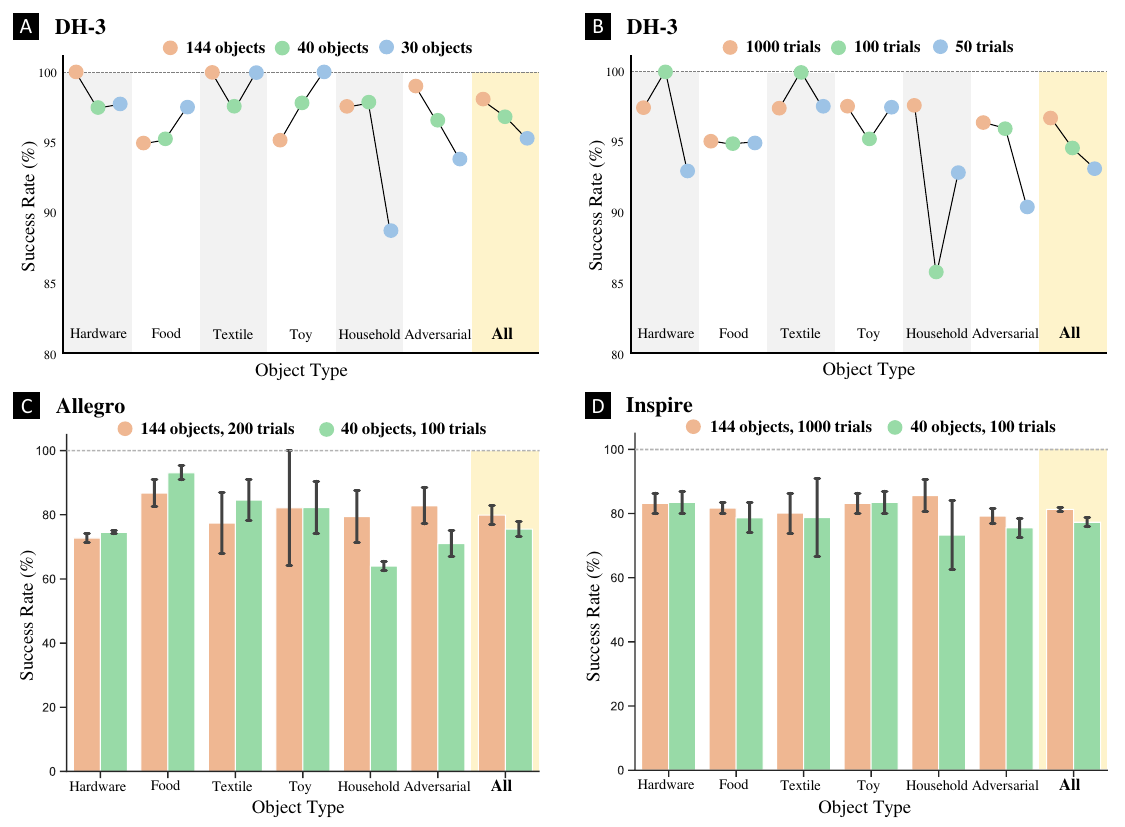}
    \caption{\textbf{Success rates on the testing set after training on reduced real-world data.} \textbf{(A)}: We reduce the training object number from 144 to 40 and 30 respectively and test the success rates on different categories of objects. \textbf{(B)}:  With 40 training objects, we reduce the data from around 1000 trials per grasp type to 100 trials and 50 trials respectively. \textbf{(C) and (D)}: When reducing the training data on fewer training objects and fewer grasp attempts, success rates on both Allegro hand and Inspire hand only decrease slightly, showing good generalization ability and high learning efficiency of our method.}
    \label{fig:efficiency}
\end{figure}

\subsection{Reducing Real-World Training Burden}
In previous experiments, we collected a considerable volume of real-world training data, approximately 1,000 trials per grasp type, on the full set of 144 training objects. In this section, we aim to alleviate the demands of real-world training by assessing the model's performance under reduced object and trial conditions. For the sake of simplicity, our evaluation concentrates on the 3-finger hand in this section.

We initiate this exploration by reducing the number of objects utilized in real-world trial-and-error attempts. Two smaller sets consisting of 40 and 30 objects were adopted for our experiments. The details of the training object sets are given in Materials and Method. To ensure a fair comparison with the original object set, we collect an equivalent number of around 1,000 grasp samples for each grasp type, ensuring the convergence of the grasp decision model.

Figure~\ref{fig:efficiency}A presents the experimental results. As the training object count reduces from 144 to 40, the grasp success rate achieves 96.7\% on all testing objects, representing only a marginal decrease of 1.1\%. Such subtle performance degradation, given a nearly 3/4 reduction in training objects, showcases the robustness of our approach.  A further reduction to 30 training objects results in an overall grasp success of 95.1\%. This translates to a further decrease of 1.6\%, yet the performance remains notably promising. Our experiments demonstrate that, with proper learning methods, the thousands of training objects adopted in previous systems\cite{dexnet4, wang2023dexgraspnet} are not necessary.

Since the performance degradation when reducing the object set from 144 to 40 and from 40 to 30 is comparable, while reducing the training object set from 40 to 30 does not significantly lower the training burden, we opt to proceed with the 40-object training set for subsequent experiments.

We then explore the impact of reducing the number of trials and errors for each grasp type. On the 40 training objects, we reduce trials and errors from approximately 1,000 attempts per grasp type to 100 attempts and 50 attempts, respectively. Figure~\ref{fig:efficiency}B presents the real robot testing results. When training with 100 trials per grasp type, the success rate reaches 94.5\% on average on all objects. We show the whole grasping process in Movie \hyperlink{movie_s7}{S7}. This success rate is strikingly high given the limited number of real-world training samples. Previous literature~\cite{xu2023unidexgrasp,liu2023dexrepnet} often required millions of grasp attempts in simulation to achieve grasping proficiency. Further reducing the trials to 50 attempts per grasp type yields a success rate of 93.1\% on all objects. These results demonstrate the high learning efficiency of our method, which requires only a small number of grasp attempts for convergence. In our following experiments, given the already high efficiency of 100 trials per grasp type, we adopt this setting for learning.

\subsection{Dexterous Grasp Learning with 40 Objects and 100 Attempts}
In the previous section, we demonstrated the robust grasping policy acquired by the 3-finger gripper through a significantly limited amount of training data and real-world attempts. In this section, we extend the validation of such a learning paradigm to the other two robotic hands utilized in this study.

We directly assess the performance of training using 40 objects with 100 trials for each grasp type on the four-finger and five-finger hands. Depending on the number of grasp types, the total real-world training samples amount to 1,000 and 800 for these two robotic hands, respectively. This significantly reduced volume of real-world training samples, nearly 1/10 of the original experiments, presents a territory in grasp learning that is unexplored by previous work.

Figure~\ref{fig:efficiency}C and Figure~\ref{fig:efficiency}D display the detailed success rates of real-world experiments. The average success rates stand at 75\%, and 77\% for all objects. The grasping process is recorded in Movies \hyperlink{movie_s8}{S8}, \hyperlink{movie_s9}{S9}, \hyperlink{movie_s10}{S10} and \hyperlink{movie_s11}{S11}. It's striking that the success rates show minimal decreases compared to the original performance. This observation demonstrates the substantial learning efficiency enabled by our methodology. Such proficiency allows diverse robotic hands to acquire dexterous grasping ability in real-world settings. 

Notably, this efficiency surpasses that observed in human infants, who typically require months of practice to develop visually guided grasping skills. The grasp success rates for human infants reach 61.9\% at 8 months old~\cite{domellof2015infant}, which involves thousands of practice attempts starting at 4 months old~\cite{newell1989task}. It is noteworthy that our grasping results are achieved based solely on visual perception, with no tactile feedback.

\subsection{Influence of Grasp Types}

\begin{figure}
    \centering
    \includegraphics[width=\textwidth]{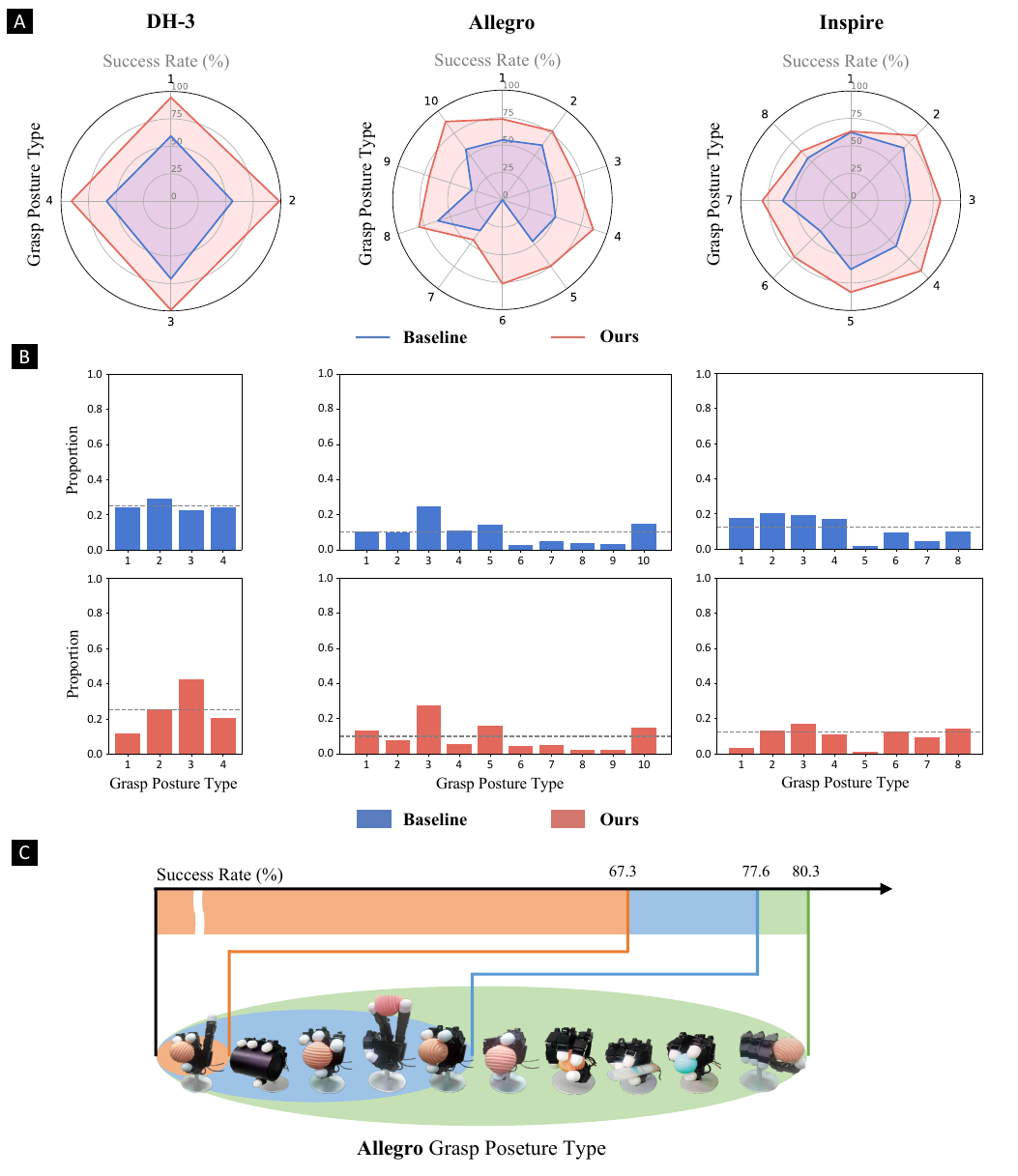}
    \caption{\textbf{Analysis of the influence by grasp type.} \textbf{(A)} A breakdown analysis of grasp success rates on different grasp types for each robotic hand. \textbf{(B)} The selected frequency of different grasp types for each robotic hand during testing. \textbf{(C)} Grasping success rates when using different portions of grasp types for the allegro hand.}
    \label{fig:grasp_type}
\end{figure}

\subsubsection*{Accuracy of Different Grasp Poses}

In the above experiments, we have shown that our method can enable efficient grasp learning with high success rates. Here we further analyze the success rates of each robotic hand with a detailed breakdown according to their respective grasp types. For clarity purposes, we number each grasp type, as illustrated in Figure~\ref{fig:index}. The results trained on 40 objects and 100 grasp attempts per grasp type are adopted for analysis, as depicted in Figure~\ref{fig:grasp_type}A. For each hand, we can see that different grasp types have different difficulties in dealing with grasping. Usually, the success rates after learning are dramatically higher than the baseline method. However, there also exist some exceptions. For example, the grasp type 1 of the five-finger Inspire hand after training yields a close success rate to the baseline. After inspection, we found that this grasp type was selected fewer times after the training. We anticipated that other grasp types might be more confident to grasp if the objects can be grasped by multiple types, which leaves some hard cases for this grasp type.

\subsubsection*{Distribution of Grasp Types}
A natural question that arises is whether the system learned by our method can demonstrate a variety of grasp types. From the example above, it is possible that the system may achieve high success rates by favoring one or two grasp types while ignoring diversity. To address this, we analyze how frequently each grasp type is selected during testing to verify whether our system indeed learns diverse grasp poses. To quantify the frequency of each grasp type, we normalize by dividing the number of grasp attempts for each type by the total number of grasp attempts across all types.

To establish a baseline, we examine the frequency of grasp types obtained by the baseline method, reflecting the inherent frequency determined solely by collision detection. Grasp types prone to collision with the scene naturally constitute a smaller fraction among all types. This baseline grasp type frequency serves as a reference for natural distribution. The top row of Figure~\ref{fig:grasp_type}B illustrates the grasp type frequency for different robotic hands. Notably, the three-finger hand exhibits a balanced distribution, whereas the four- and five-finger hands display more unbalanced distributions. This discrepancy arises from the fact that the fingertips of the three-finger hand consistently point in the same direction along the approach vector, resulting in similar collision situations across different types. Conversely, the four- and five-finger hands exhibit types with greater variance, including some that are prone to collision with the scene.

Then, we present the frequency of grasp pose after employing our learned system. The statistics are given in the second row of Figure~\ref{fig:grasp_type}B. For the three-finger hand, type 3 presents an increasing ratio among all grasp types. The reason is that this grasp type presents a higher success rate, and usually has a higher grasp quality score than other grasp types. However, the other three grasp types are also frequently selected. For the four- and five-finger hand, the grasp frequency is similar to the baseline method. These results affirm that our learned system adeptly captures diverse grasp poses, achieving high success rates without compromising grasp diversity.

\subsubsection*{Reducing Grasp Types}
Another question for multi-finger grasping is whether employing multiple grasp types is necessary, given the argument that a single power grasp might be sufficient for good results. However, we argue that incorporating multiple types enhances flexibility, particularly when faced with cluttered scenarios. To prove that, we conducted a targeted experiment to compare grasping outcomes with varying numbers of grasp types. Specifically, we employed the best grasp model trained with 144 objects for the Allegro hand, initially defined with 10 grasp types. In our experiment, we compared the original model with two modified versions that use fewer grasp types. The first version was limited to a single grasp type, specifically the one that achieved the highest overall success rate across all types. The second version used a subset of the five most effective grasp types, chosen based on their individual success rates. For simplicity, the evaluation focused exclusively on adversarial objects due to the performance similarity with that on the entire object set. The resulting success rates are detailed in Figure~\ref{fig:grasp_type}C.

The original method, employing 10 grasp types, achieved an 80.3\% success rate on the test set. In contrast, utilizing only a single grasp type led to a reduction in the grasp success rate to 67.3\%. Employing five grasp types performed better, resulting in a success rate of 77.6\%, but is still inferior to the original method. Our experimental results show that increasing the number of grasp types can improve overall grasp success rates. One reason for this improvement is that a greater variety of grasp types provides more flexibility, enabling the hand to better adapt to different object shapes, sizes, and orientations. Additionally, using multiple grasp types can increase tolerance for collisions, allowing the hand to adjust its grasping strategy based on spatial constraints, particularly in cluttered environments. It is noteworthy that, on the other hand, our results also reveal that the benefits derived from further adding grasp types would eventually saturate. Thus, it is reasonable to adopt diverse yet limited grasp types, which optimizes both grasp success rates and learning efficiency.

\section{Discussion}

\subsection{Efficiency Analysis}
It is surprising to see that our grasp system can be learned for different hands so efficiently. Previous work for multi-finger grasping usually requires thousands of objects or millions of grasp samples~\cite {wang2023dexgraspnet, lum2024dextrahg,
singh2025dextrahrgb}. In the deep learning era, it seems to be an underlying rule that we need to train a robot system on as many objects as possible to have good generalization ability. However, the satisfactory performance of our system breaks this intuition. What are the key factors for our method to learn so efficiently and generalize so well? There are two aspects of learning efficiency in our system, the first is that we only need 40 objects and the second is that we only need hundreds of trials for each hand. Here we discuss how our method achieves efficiency in these two aspects.

\subsubsection*{Representation Model}
Representation plays a crucial role in our system for real-world learning, as it must map different geometries into a contact-centric grasp representation. How can the system, trained on only 40 objects, generalize to hundreds of unseen objects? We address this by conducting a geometry coverage analysis, revealing that \textbf{scaling up data along the right dimension} is key to improving the model's generalization ability.

Our representation model takes a scene point cloud as input and outputs contact-centric grasp representations (CGRs). To train this model, we need a dataset containing scene point clouds with annotated CGRs across various geometries. Since CGRs depend on local geometry, a representative dataset must include diverse local geometries to effectively train deep networks. While many researchers intuitively attempt to collect more training objects to achieve this, our geometry coverage analysis demonstrates that more objects do not necessarily lead to richer local geometries.

We begin by defining the local geometry used in our analysis. Specifically, the representation network operates in a partial observation scenario, where it infers contact positions and normals on unobserved surfaces based on the observed geometry. Although each normal and contact point is predicted independently, we consider the minimal continuous components of local patches, enclosed by the simplest form of grasping—an antipodal grasp—as a foundational element in our analysis for consistency.

\begin{figure}[t]
    \centering
    \includegraphics[width=1\textwidth]{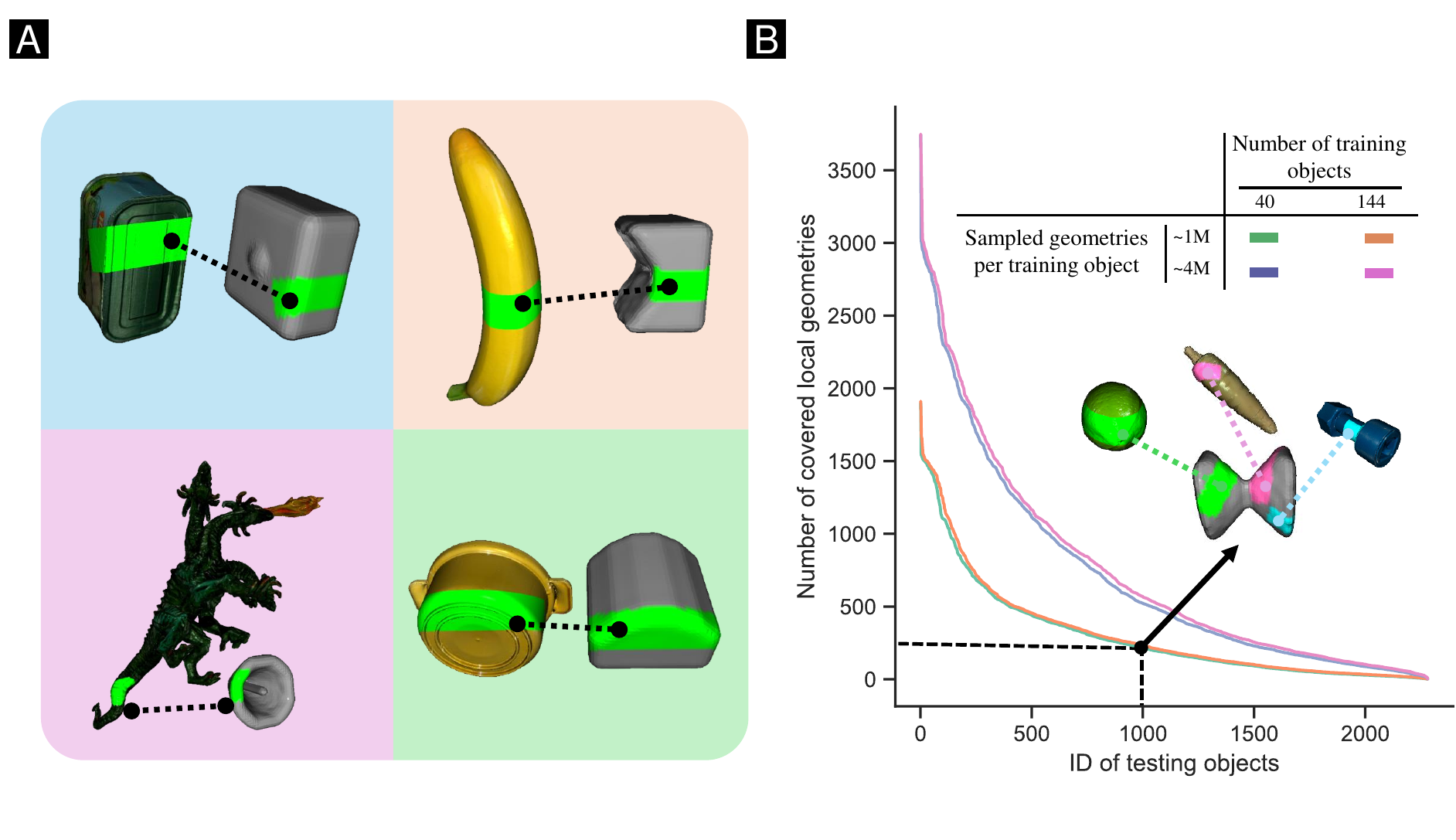}
    \caption{\textbf{Geometry coverage analysis.} \textbf{(A)}: The colored objects are our training objects and the gray objects are in the EGAD! object set. The surfaces highlighted in green and connected by a dotted line have similar local geometries. We see that although the training object and testing object have very different overall shapes, we can find local geometries on them that are pretty similar. \textbf{(B)}: The local geometry coverage curves on the testing set given different choices of scaling up the training set. The $x$-axis denotes the ID of each testing object, and the $y$-axis denotes the number of covered local geometries on each testing object. An example is given where the 1000-th testing object has around 250 covered local geometries. We only draw 3 of them for illustration.
    }
    \label{fig:coverage}
\end{figure}

Next, we define geometry coverage in the analysis. Given a training and testing object set, a local geometry on a test object is considered "covered" if it closely resembles a local geometry from the training object set. We define similarity by a chamfer distance smaller than 1mm, with examples illustrated in Figure~\ref{fig:coverage}A. For a training dataset, we can assess the diversity of local geometries by counting the number of covered geometries on test objects.

In practice, when constructing a training dataset, we need to generate labels for a fixed number of local geometries selected from the training objects, constrained by computational resources. There are two possible dimensions  along which to collect more local geometries: increase the number of training objects or increase the sample density on each training object. To assess which dimension is more effective for increasing local geometry diversity, our analysis is conducted as follows. We collect two training object sets: $\mathbb{S}$, with 40 objects, and $\mathbb{L}$, with 144 objects. For each object set, we sample 1 million and 4 million local geometries from each object on average, respectively (sampling details given in supplementary material). This combination results in four different training datasets. The testing object set for the coverage analysis is the EGAD! test set, which contains over 2000 complex, program-generated objects. We sample around 400 local geometries on each test object and evaluate if they are covered. Figure~\ref{fig:coverage}B shows the number of covered geometries on each test object, considering different training sets and sampling densities. 

Surprisingly, we found that increasing the number of training objects does not significantly increase the coverage rate on the test set. However, increasing the sampling density of local geometries per object leads to a dramatic increase in coverage—\textit{even when the total number of sampled geometries is similar to increasing the object count}. This result demonstrates that increasing sample density for each training object is far more impactful than increasing the number of objects.

Based on this analysis, we prioritize scaling up the label density of CGRs on each training object, rather than increasing the number of training objects, when constructing our dataset. By training on over a billion CGRs, our model has learned to map local geometries to grasp representations effectively, thereby enhancing its ability to generalize to novel objects.

\subsubsection*{Grasp Decision Model}
In the previous section, we discussed how our representation model can generalize well to novel scenes despite being trained on only 40 objects. Now, we turn to the grasp decision model and discuss why it can learn grasp success from just hundreds of trial-and-error attempts. Here, we highlight a few possible reasons.

First, the representation captures all the relevant information about force closure that can be extracted from vision. For a point-to-plane contact problem, the force-closure condition must satisfy the following criteria~\cite{dai2018synthesis}:
\begin{equation}
\begin{aligned}
     &Gf = 0, \\
     &GG^\top > \epsilon I_{6\times6},\\
     &f_i^\top n_i > \frac{1}{\sqrt{\mu^2+1}}|f_i|,
\label{eqn:forceclosure}
\end{aligned}
\end{equation}
where $f$ is the vector of contact forces acting at each contact point, $G$ is the grasp matrix determined by the positions of the contact points, and $n_i$ represents the surface normal at the $i$-th contact point. The latter two parameters are the only aspects that a vision model can estimate, and are generated by our representation model. The grasp decision model needs only to learn whether the forces $f$ exerted by the gripper for different grasp types can satisfy Equation~\eqref{eqn:forceclosure}, given a friction coefficient $\mu$. Although the friction coefficient is unknown, the model tends to learn an average behavior from the training set.

Second, the representation is compact. Instead of dealing with high-dimensional data like images or point clouds, we reduce the input to a 1D vector that represents the shape. This compactness simplifies the mapping from input to grasp quality, making it easier for the grasp decision model to learn.

\subsection{Method Positioning}

The evolution of visually guided dexterous grasping methodologies within robotics has developed two prominent paradigms: the 6D pose estimation paradigm and the end-to-end grasp learning paradigm. The former relies on the precise estimation of an object's 6D pose and then calculates the hand pose accordingly. It can transfer across different robotic hands easily, but requires prior knowledge of the object's model. On the other hand, the end-to-end grasp learning models do not require explicit object knowledge, yet the trained models lack transferability across different robotic hands.

Our proposed approach explores a middle ground between these two paradigms, which ombining the advantages of both. By developing a contact-centric grasp representation that encapsulates the scene's contact information, we eliminate the need for an object's model beforehand. The CGR preserves critical information pertinent to grasp quality, endowing our system with adaptability and applicability across different morphologies of robotic hands. Moreover, by eliminating the need for an accurate kinematic model, which was frequently used in previous work learned in simulation~\cite{xu2023unidexgrasp,wan2023unidexgrasp++}, our method is suitable for soft hand grasp learning.

\subsection{Integration with Tactile Sensors}
Future directions for research entail expanding the scope of the contact-centric grasp representation model to include a wider array of tactile and sensory information, enabling a more comprehensive understanding of object manipulation. Tactile sensors encapsulate rich information concerning contact positions and contact point normals, mirroring the fundamental attributes of our representation model. This alignment highlights the potential for our method to work well with tactile sensing technology. By leveraging this alignment, the incorporation of tactile sensors can augment the current representation, further refining the contact-centric information available for decision-making in our learning paradigm.

\section{Materials and Method}

\subsection{Baseline Method}
Here we introduce our baseline method of multi-finger grasping. Currently, our community can achieve human-level robotic grasping with a parallel-jaw gripper~\cite{fang2023anygrasp}. An intuitive approach for multi-finger grasping is to mimic the behavior of parallel grasping. Thus, we propose a baseline method that discovers the principal closing axis of a robotic hand and aligns it with a parallel grasp pose. First, for each grasp type of a robotic hand, we manually designate its principal closing axis, which is the primary direction along which the fingers converge when the hand closes to grasp an object. Then, given a parallel grasp pose and a grasp type of a hand, we can align the multi-finger hand's principal closing axis to the parallel grasp pose. Previous literature~\cite{fan2019optimization,fan2018real} also explored similar ways to initialize a multi-finger grasp. In Figure~\ref{fig:principal} we illustrate the alignment example. 

When grasping with a selected robotic hand, we first generate multiple high-score antipodal grasp poses for a single-view point cloud using the AnyGrasp library~\cite{fang2023anygrasp}. Then, for each antipodal grasp pose, we align the robotic hand configured in all grasp types with the antipodal pose. It means that for each antipodal grasp pose, we would have multiple multi-finger grasp candidates with different types. For all of the multi-finger grasp candidates across the scene, we run a collision detection based on the partial-view point cloud and robotic hand model. We select the grasp type assigned with the highest antipodal grasp score for the remaining grasp candidates without collision. If multiple grasp candidates have the same grasp score, we randomly select one as the final grasp pose.
 
\subsection{Algorithm Details}
Next, we introduce the details of our algorithm, which consists of three steps: learning the representation model, mapping from representation to grasp pose, and learning the grasp decision model.
\subsubsection*{Representation Model} \label{repre_model}
Our representation model takes a partial-view point cloud as input and generates the contact-centric grasp representation $r$ for different rotation $\mathbf{R}_{3d}$ and translation $\mathbf{t}_{3d}$ across the scene. It may seem initially challenging to establish the representation model, given that this representation demands full surface information, and the $r$ needs to be predicted for SE(3) space across the scene. However, recent advancements in grasp pose detection algorithms have successfully learned the mapping from partial-view point clouds to antipodal grasp poses across the scene, unveiling the feasibility of learning the mapping from partial-view point clouds to the proposed intermediate representation. Specifically, prior works, such as graspnet-baseline~\cite{graspnet} and GSNet~\cite{wang2021graspness}, have predicted the gripper opening widths and antipodal scores for discretized rotation $\mathbf{R}_{3d}$ and translation $\mathbf{t}_{3d}$ across the scene:   
\begin{align}
s = \bigg\{ & (w_{\alpha_i}, \mu_{\alpha_i})_j  \bigg|  \alpha_i = 0, \frac{2\pi}{N}, \ldots, \pi-\frac{2\pi}{N}, \, j = 1, 2, \ldots, M;  \mathbf{R_{3d}}, \mathbf{t_{3d}} \bigg\},
\label{grasprep}
\end{align}  
where $w_{\alpha_i} = 2\times\max(d_{\alpha_i}, d_{\alpha_i+\pi})$ and  $\mu_{\alpha_i} = \max(\tan(\theta_{\alpha_i}), \tan(\theta_{\alpha_i+\pi}))$ are the gripper opening width and antipodal grasp quality metric defined in~\cite{graspnet}. This representation shares a structural resemblance with our contact-centric grasp representation in Equation~\eqref{eqn_r3d}. Thus, we opt to build our representation model upon the GSNet~\cite{wang2021graspness} architecture.
 
When predicting the representation, it is intractable to account for every possible \(\mathbf{R}_{3d}\) and \(\mathbf{t}_{3d}\) in continuous space. Previous work~\cite{graspnet,wang2021graspness} addressed this by selecting orientation \(\mathbf{R}_{3d}\) from 300 discretized directions, voxelizing the scene, and selecting only \(\mathbf{t}_{3d}\) that lies on object surfaces. However, the total number of resulting combinations still remains quite large. In~\cite{wang2021graspness}, a metric called “graspness” is proposed as a heuristic to bias sampling towards \(\mathbf{t}_{3d}\) and \(\mathbf{R}_{3d}\) values that have a higher probability of generating successful grasp poses. This metric includes two components: “point-wise graspness” and “view-wise graspness.” Point-wise graspness is calculated by counting the ratio of high-score antipodal grasp poses among all poses at a given \(\mathbf{t}_{3d}\), while view-wise graspness counts this ratio among all grasp poses at a given \(\mathbf{R}_{3d}\) for a sampled \(\mathbf{t}_{3d}\). These two scores are learned jointly within the grasp pose detection network and guide sampling during inference.

In this work, since we aim to train a hand-agnostic representation model, we define a new “graspness” score for each \(r\) to indicate its suitability for subsequent grasping across different robotic hands. Intuitively, for a point \((\alpha_i, d_{i}, \theta_i)\), a robotic hand achieves better contact when \(\theta_i\) is small, meaning the surface normal \(n_i\) is opposite to the contact direction (assuming the robotic finger approaches towards the polar pole of the local coordinate frame). Additionally, geometries with many high-score antipodal grasp poses tend to be easier for dexterous hands to grasp. Thus, we define “graspness” in this paper as the sum of \(\theta_i\) values below a threshold and the number of antipodal grasp poses in \(r\). This definition helps the model reduce candidates for representation prediction within a scene without significantly affecting accuracy.

Similar to GSNet, our representation model consists of three cascaded modules. Firstly, a Minkowski Engine~\cite{choy20194d} backbone takes the single-view point cloud as input, encodes their geometric features, and outputs a computed feature vector for each input point. Then a multi-layer-perception (MLP) takes the features of each point and generates a point-wise graspness heatmap. We sample 1024 seed points with high graspness, and forward these points to another MLP block. It outputs the view-wise graspness scores for 300 approach directions towards each seed point respectively. We then select the direction with the highest graspness score for each point, group the features with cylinder grouping~\cite{fang2020graspnet} along that direction and forward the grouped features for each point through a final MLP block. This final layer outputs $r$ for $N=48$ in-plane rotations and $M=5$ grasp depths, which are 0.005m, 0.01m, 0.02m, 0.03m and 0.04m respectively.

\subsubsection*{Mapping from Representation to Multi-Finger Grasp Candidates}
After we obtain the representation $r$ at different positions across the scene, we link them with different grasp types of a robotic hand to generate multi-finger grasp candidates. In theory, since we have predicted the contact information, we can already generate suitable multi-finger grasp candidates through optimization~\cite{miller2004graspit,liu2021synthesizing}. However, for simplicity, we follow the same technique adopted in the baseline method to generate multi-finger grasp candidates. Such a design also facilitates fair comparison with the baseline method and shows how our grasp decision network improves the grasping ability.

For each predicted CGR with the form of Equation~\eqref{eqn_r3d}, we calculate the corresponding antipodal grasp representation defined in Equation~\eqref{grasprep}. Then the CGRs with top-500 antipodal grasp scores are selected. These representations are associated with different multi-finger grasp candidates following the same procedure of the baseline method. After this process, we query the orientation $\mathbf{R}_g$ and translation $\mathbf{t}_g$ of the multi-finger grasp candidates associated with the CGRs (more details in supplementary material). Together with the associated grasp types, we map the CGRs to multi-finger grasp candidates.

\subsubsection*{Learning Multi-Finger Grasping} \label{model_architecture}
For each sampled grasp candidate $g_i$, we learn a mapping from its corresponding CGR $r_i$ to grasp success probability. This mapping is approximated through the grasp decision model, using training data collected via trial and error: 
\begin{equation}
\nonumber\alpha = \Psi(r_i, g_i, h).
\end{equation}

In practice, we  train different decision models for different robotic hands, denoted as  $\Psi_h(r_i, g_i)$. Since the grasp types of each hand are discretized, we further decompose the classification of different grasp types into different sub-models:
\[
\nonumber\Psi_h(r_i, g_i) = \sum_{\mathbf{q}} \mathbb{I}(g_i, \mathbf{q}) \, \Psi_{h,\mathbf{q}}(r_i, \mathbf{R}_g, \mathbf{t}_g),
\]
where $\mathbb{I}(g_i, \mathbf{q})$ is an indicator function that is 1 when $g_i$ matches the grasp type $\mathbf{q}$ and 0 otherwise. Since $\mathbf{R}_g$ and \(\mathbf{t}_g\) are functions of \(r_i\), we can simplify the input to the sub-models by removing \(\mathbf{R}_g\) and \(\mathbf{t}_g\). Thus, \(\Psi_{h,\mathbf{q}}(r_i, \mathbf{R}_g, \mathbf{t}_g)\) can be reformulated as \(\Psi_{h,\mathbf{q}}(r_i)\), where the computation of \(\mathbf{R}_g\) and \(\mathbf{t}_g\) is implicit in the model. We empirically found that using different sub-models for different grasp types gives better performance. The input to the model consists of the CGR of a selected grasp candidate. The model's output is a score of whether the selected grasp would be successful. Details of the model is given in supplementary material. For simplicity, we regard the combination of all sub-models for each robotic hand as a single model and still refers to it as the grasp decision model.

\subsubsection*{Detection Post-Processing}
\label{detection post processing}
Following the grasp decision model's output, we select grasp poses with high-quality scores, typically exceeding 0.9. Collision detection is then performed by voxelizing the pre-shaped multi-finger hand and examining intersections between the hand voxels and the scene point cloud using the Open3D library. The final grasp pose is chosen from those grasp poses without collision with the scene, with the highest grasp quality score.

\subsection{Training Environment}

\definecolor{darkgrey}{rgb}{0.65, 0.65, 0.65}
\begin{algorithm}[!t]
\caption{Multi-finger Grasping Data Collection for Robotic Hand $h$}\label{alg:anyhand-learn}

\textbf{Input:} the expected size $K$ of the grasp dataset.

\textbf{Output:} the collected grasp dataset $\mathbf{G}$.

\begin{algorithmic}[1]
\State $\mathbf{G} \leftarrow \emptyset$
\While{$|\mathbf{G}| < K$}
\State The \textbf{\textit{robot}} moves to the ready pose
\State $\mathcal{P} \leftarrow \textbf{\textit{camera}}.\text{perception}$ \Comment{\textcolor{darkgrey}{capture RGBD images and transform into point cloud}}
\State $\mathcal{R} \leftarrow \Phi (\mathcal{P})$ \Comment{\textcolor{darkgrey}{generate scene representation from the point cloud}} 
\State Sample a CGR $r \in \mathcal{R}$ in the scene
\State Sample a grasp type $\mathbf{q} \in \{\mathbf{q}_1,...,\mathbf{q}_c\}$
\State $[\mathbf{R}_g\ \mathbf{t}_g]\leftarrow\text{compute\_grasp\_pose}(r)$
\Comment{\textcolor{darkgrey}{Map CGR to the grasp pose}}
\If{$\text{collision\_detection}([\mathbf{R}_g\ \mathbf{t}_g\ \mathbf{q}], \mathcal{P}; h)$} \\\Comment{\textcolor{darkgrey}{Check if the multi-finger grasp pose will collide with the scene point cloud}}
\State \textbf{continue}
\EndIf
\State The \textbf{\textit{robotic hand}} executes the multi-finger grasp pose $[\mathbf{R}\ \mathbf{t}\ \mathbf{q}]$
\State Record the grasp result $S$ \Comment{\textcolor{darkgrey}{Collect trial-and-error results}}
\State $\mathbf{G} \leftarrow \mathbf{G} \cup \{\left<r, \mathbf{R}_g, \mathbf{t}_g, \mathbf{q}, S\right>\}$
\EndWhile \\
\Return the collected grasp dataset $\mathbf{G}$
\end{algorithmic}
\label{alg_datacollection}
\end{algorithm}

\subsubsection*{Training Object Set}
Our experiments involve three different training object sets, the larger dataset $\mathbb{L}$ contains 144 objects, the smaller one $\mathbb{S}$ contains 40 objects, and the tiniest one $\mathbb{T}$ contains 30 objects. $\mathbb{L}$ is the training set collected in AnyGrasp~\cite{fang2023anygrasp}. $\mathbb{S}$ encompasses the 40 training objects featured in the original GraspNet-1Billion dataset, and $\mathbb{T}$ includes 30 randomly selected objects from $\mathbb{L}$. In Figure~\ref{fig:train_obj} we detail the three training object sets.

\subsubsection*{Data Annotation and Collection}

To facilitate the training of our representation model, we re-annotate the GraspNet-1Billion dataset. The training set consists of 100 scenes made up of 40 objects. Each scene includes 256 RGBD images, each of which can be transformed into a single-view point cloud. Instead of the original antipodal grasp representation (illustrated in Equation~\eqref{grasprep}), we annotate the contact-centric grasp representation as per Equation~\eqref{eqn_r3d} for the 100 training scenes.

Our process begins by voxelizing the 3D mesh of each training object with a resolution of 0.005 m. We collect all points on the voxelized object surface, denoted as \(\{\mathbf{t}_{3d}^{(i)}\}\), where \(i\) indexes each individual surface point. For each surface point \(\mathbf{t}_{3d}^{(i)}\), we sample 300 approach directions \(\{\mathbf{R}_{3d}^{(j)}\}\), where \(j\) indexes the sampled directions. We then compute the CGR \(r\) for each combination of \(\mathbf{t}_{3d}^{(i)}\) and \(\mathbf{R}_{3d}^{(j)}\). This computation relies on the complete mesh of the object. The computed CGRs are then projected from each training object to the training scenes based on the object’s 6D pose provided in the original dataset.

After generating the CGRs for each scene, we apply a simple post-processing step to verify grasp feasibility. For each CGR, we check whether a cylindrical region extending backward along the approach direction collides with the tabletop or other objects in the scene. If a collision is detected, we set the CGR to a zero vector, indicating it is not a viable grasp candidate. This post-processing step helps reduce the likelihood of robotic hand collisions within the scene.

To train the grasp decision model, we collect grasping data by trial and error. Previously, most of the grasp attempts related to multi-finger grasping were collected within a simulation environment. Nevertheless, significant gaps may arise due to the inherent differences between the simulation and real environments. Thus, in this paper, we directly collect grasping data in a real-world environment.

We provide an overview of the complete data collection pipeline, summarized in Algorithm~\ref{alg_datacollection}. Initially, we randomly place objects on the table. We then run the representation model to generate dense contact-centric grasp representations for the scene. We sample a CGR and a grasp type of the robotic hand, and map the CGR to a multi-finger grasp pose. Collision detection is performed to ensure that the grasp pose does not collide with the scene. If no collision happens, we execute the grasp process. During this process, we record whether the grasp is successful and store the necessary information in the dataset.

\subsubsection*{Training Details}
For the representation model, the input point clouds are down-sampled with a voxel size of 0.005m. In practice, we set the parameters of the 3D representation $N$ and $M$ in Equation~\eqref{eqn_r3d} to 48 and 5 respectively. The model is trained on the re-annotated GraspNet-1Billion dataset using one Nvidia A100 GPU with Adam optimizer~\cite{kingma2014adam} and an initial learning rate of 0.001. The learning rate follows a descent strategy and we adopt ``poly'' policy with $power=0.9$ for learning rate decay. The model is trained from scratch with a batch size of 4. For data augmentation, we randomly flip the scene horizontally and randomly rotate the points by Uniform$[-30^\circ,30^\circ]$ around the $z$-axis (in the camera coordinate frame). We also randomly translate the points by Uniform$[-0.2\text{m},0.2\text{m}]$ in the $x$- or $y$-axis and Uniform$[-0.1\text{m},0.2\text{m}]$ in the $z$-axis.

For the grasp decision model, since we have a relatively limited amount of collected data, our model is trained for only 20 epochs to avoid overfitting. We leverage the Adam optimizer~\cite{kingma2014adam}. The learning rate follows a segmented descent strategy starting from 0.0001, and the batch size $\mathbf{Z}$ is set to 128 to optimize training efficiency. Since the network is quite small, we train the model on a laptop with NVIDIA 1650 GPU.

\subsection{Experimental Procedure}
In each experiment, we randomly distribute objects from different categories in the robot workspace. During the grasping process, the partial-view point cloud captured by the camera is fed into our representation model. When collecting training data, we follow the procedure in Algorithm 1. During testing, we first choose 100 CGRs from the outcome of the representation model, which has the top-100 antipodal grasp scores. These CGRs are mapped to multi-finger grasp candidates, and given the number of predefined types for each robotic hand, the total number of multi-finger grasp candidates varies (\textit{e.g.}, we define 4 grasp types for the three-finger hand, thus it has 400 grasp candidates). These grasp candidates are fed into our grasp decision model. The grasp candidates with the top-200 grasp quality scores then undergo collision detection post-processing. The grasp pose that passes collision detection and has the highest grasp score is selected as the final multi-finger grasp pose in the camera's coordinate system. It is subsequently converted into the world coordinate system and sent to the UR5 robot through socket communication. The UR5's embedded motion planner navigates it to the grasp pose, where we set a waypoint 10 cm backward from the final grasp along the approach direction to avoid collision during movement. Simultaneously, the robotic hand is configured to the selected grasp type. After the robot arm reaches the target pose, the robotic hand closes the fingers until the grasping force reaches a predefined limit. The robot then lifts the object and moves it to the top of the bin and drops the object. The experiment concludes with manually recording whether the robotic hand successfully move the object to target position.

\section*{Acknowledgments}
The authors would like to thank Xiaolin Fang for the helpful revision and Antonia Bronars and Jiang Zou for the helpful discussion. The latex template is borrowed from HIL-SERL.


\newpage
\balance

\bibliography{main}

\newpage

\appendix

\section*{\Large \textbf{Supplementary Methods}}

\subsection*{Query 6D Grasp Pose for CGR }
When we map a CGR to grasp pose, we first calculate the antipodal grasp representation of the CGR. Given a CGR
\begin{align}
r = \bigg\{ & (d_{\alpha_i}, \theta_{\alpha_i})_j \, \bigg| \, \alpha_i = 0, \frac{2\pi}{N}, \ldots, 2\pi-\frac{2\pi}{N}, \, j = 1, 2, \ldots, M; \mathbf{R}_{3d},\mathbf{t}_{3d} \bigg\},
\end{align}
the antipodal grasp representation is calculated by
\begin{align}
\nonumber s = \bigg\{ & (w_{\alpha_i}, \mu_{\alpha_i})_j  \bigg|  \alpha_i = 0, \frac{2\pi}{N}, \ldots, \pi-\frac{2\pi}{N}, \, j = 1, 2, \ldots, M;  \mathbf{R_{3d}}, \mathbf{t_{3d}} \bigg\},
\label{grasprep_sup}
\end{align}  
where  $w_{\alpha_i} = 2\times\max(d_{\alpha_i}, d_{\alpha_i+\pi})$ and  $\mu_{\alpha_i} = \max(\tan(\theta_{\alpha_i}), \tan(\theta_{\alpha_i+\pi}))$. After we obtained $s$, we choose the $\alpha_i$ and $j$ that has the maximum antipodal grasp score:
\[
(\alpha_i^*, j^*) = \underset{\alpha_i, j}{\operatorname{arg\,max}} \, (\mu_{\alpha_i})_j.
\]
We add the rotation $\alpha_i^*$ and translation corresponds to the $j^*$-th section along approach direction to $\mathbf{R_{3d}}$ and $\mathbf{t_{3d}}$:
\begin{align}
\nonumber \mathbf{R_{g}} &= \mathbf{R_{3d}} \cdot \mathbf{R}_z(\alpha_i^*),\\
\nonumber \mathbf{t_{g}} &= \mathbf{t_{3d}} + \mathbf{d}(j^*) \cdot \mathbf{R_{3d}} \cdot \mathbf{R}_z(\alpha_i^*) \cdot \mathbf{z},
\end{align}
where:
\[
\mathbf{R}_z(\alpha_i^*) = \begin{bmatrix} \cos(\alpha_i^*) & -\sin(\alpha_i^*) & 0 \\ \sin(\alpha_i^*) & \cos(\alpha_i^*) & 0 \\ 0 & 0 & 1 \end{bmatrix}, 
\]
$\mathbf{d}(\cdot)$ is a function that maps the index $j$ of the section to its actual depth along the approach direction (maps $\{1,2,3,4,5\}$ to $\{0.005\text{m}, 0.01\text{m}, 0.02\text{m}, 0.03\text{m}, 0.04\text{m}\}$), and:
\[
\mathbf{z} = \begin{bmatrix} 0 \\ 0 \\ 1 \end{bmatrix}.
\]
The updated rotation $\mathbf{R_{g}}$ and translation $\mathbf{t_{g}}$ are the 6D grasp pose that corresponds to this CGR. 

\subsection*{Details of Grasp Decision Model}
Each grasp decision sub-model is learned by a neural network. It takes a contact-centric grasp representation as input and outputs a score ranging from 0 to 1 to indicate whether the corresponding grasp candidate would be successful. The input size is $2\times5\times48=480$, which is composed of distances and normal angles on 5 sections along 48 in-plane rotations.

The network comprises seven fully connected layers with a skip connection for improving robustness. Each intermediate layer consists of a fully connected layer with 1024 neurons, a batch normalization layer, and a ReLU activation function. The output of the second intermediate layer is also forwarded to the fifth intermediate layer with a skip connection. Networks for different grasp types are trained separately. We employ a loss function defined as: 
\begin{equation}\label{loss}
L = -\frac{1}{\mathbf{Z}}\sum_{z=1}^\mathbf{Z}y_{z}\log(p_{z}).
\end{equation}
In this equation, $L$ is the loss, $y$ denotes the binary label of whether the real robot trial and error succeeded or not, and $p$ represents the predicted grasp success probability by the network. $\mathbf{Z}$ denotes the batch size.

\subsection*{Local Geometry Sampling for Grasp Coverage Analysis}
The local geometries are cropped using 3D boxes defined by valid antipodal grasp poses. To obtain the grasp candidates, each object is voxel-downsampled to get grasp points in uniform distributions. $V$ approach directions are sampled on the grasp point. $A$ inplane rotation angles are sampled uniformly for each direction. On the training objects, we set $V$=100 and $A$=12 for the dense set, and $V$=50 and $A$=6 for the sparse set, respectively. In these two cases, the average numbers of local geometries for each training object are around 1M and 4M. For testing objects in the EGAD dataset, we set $V$=100 and $A$=12.

\section*{\Large \textbf{Supplementary Text}}

\subsection*{Training Object Collection}
The 144 training objects are collected from supermarkets and grocery stores, which is extended from the 40 training objects collected in GraspNet-1Billion~\cite{fang2023robust}. The principle of choosing objects is that they have a roughly different shape or some local geometries from other objects, and they are chosen by authors heuristically. We provide the 3D scanned models of the objects to support reproducible research. Figure~\ref{fig:train_obj} shows an overview of the training object.

\subsection*{Grasp Types for Different Hands} The index number for each grasp type of different robotic hands is given in Figure~\ref{fig:index}.

\subsection*{Principal Closing Axis for Different Grasp Types} We illustrate the principal closing axis for different grasp types in Figure~\ref{fig:principal}. The $x$-axis (in red) in the local coordinate frame is the approach direction and the $y$-axis (in green) is the principal closing axis of the hand. The two-finger gripper (in blue) is the corresponding antipodal grasp pose for each grasp type.

\section*{\Large \textbf{Supplementary Movies}}
We believe that presenting the complete process of our robotic grasping experiments can provide valuable insights into potential improvements for the grasping system. Additionally, it is essential to demonstrate the system’s robustness, which requires running it for an extended period. Therefore, we recorded the entire grasping process, retaining all original content without cuts, but with speed adjustments to keep the video at a reasonable length. The grasping process for each robotic hand lasts over 3 hours, with the total time across all three hands exceeding 15 hours. We applied a 20x speed-up for the collision detection phase and a 2x speed-up for the grasp execution phase. Even after these adjustments, the resulting videos still exceed 6 hours in length. Consequently, we have hosted the videos on YouTube, with the links provided below:
\begin{itemize}
    \item \hypertarget{movie_s1}{Movie S1} - Grasping with 3-finger DH-3 hand on daily objects, after training on 144 objects: \\\url{https://youtu.be/GGBesshyfxk}
    \item \hypertarget{movie_s2}{Movie S2} - Grasping with 4-finger Allegro hand on daily objects, after training on 144 objects: \\\url{https://youtu.be/HkrvWm_TTGo}
    \item \hypertarget{movie_s2}{Movie S3} - Grasping with 5-finger Inspire hand on daily objects, after training on 144 objects: \\\url{https://youtu.be/3Om7G8nMJPg}
    \item \hypertarget{movie_s4}{Movie S4} - Grasping with 3-finger DH-3 hand on adversarial objects, after training on 144 objects: \\\url{https://youtu.be/GGBesshyfxk?t=1837}
    \item \hypertarget{movie_s5}{Movie S5} - Grasping with 4-finger Allegro hand on adversarial objects, after training on 144 objects: \\\url{https://youtu.be/E7i3pqxA4RM}
    \item \hypertarget{movie_s6}{Movie S6} - Grasping with 5-finger Inspire hand on adversarial objects, after training on 144 objects: \\\url{https://youtu.be/o6LQwRgu82s}
    \item \hypertarget{movie_s7}{Movie S7} - Grasping with 3-finger DH-3 hand on daily and adversarial objects, after training on 40 objects:  \url{https://youtu.be/--5wIHfPoZs}
    \item \hypertarget{movie_s8}{Movie S8} - Grasping with 4-finger Allegro hand on daily objects, after training on 40 objects: \\\url{https://youtu.be/uhaC8NORqm4}
    \item \hypertarget{movie_s9}{Movie S9} - Grasping with 4-finger Allegro hand on adversarial objects, after training on 40 objects: \\\url{https://youtu.be/5pN6BYOH4xw}
    \item \hypertarget{movie_s10}{Movie S10} - Grasping with 5-finger Inspire hand on daily objects, after training on 40 objects: \\\url{https://youtu.be/GQDLTVjXPQk} 
    \item \hypertarget{movie_s11}{Movie S11} - Grasping with 5-finger Inspire hand on adversarial objects, after training on 40 objects: \\\url{https://youtu.be/B7qc7qRw4ss} 
\end{itemize}

\clearpage

\renewcommand{\thefigure}{S\arabic{figure}}

\setcounter{figure}{0}

\begin{figure*}
    \centering
    \includegraphics[width=1\textwidth]{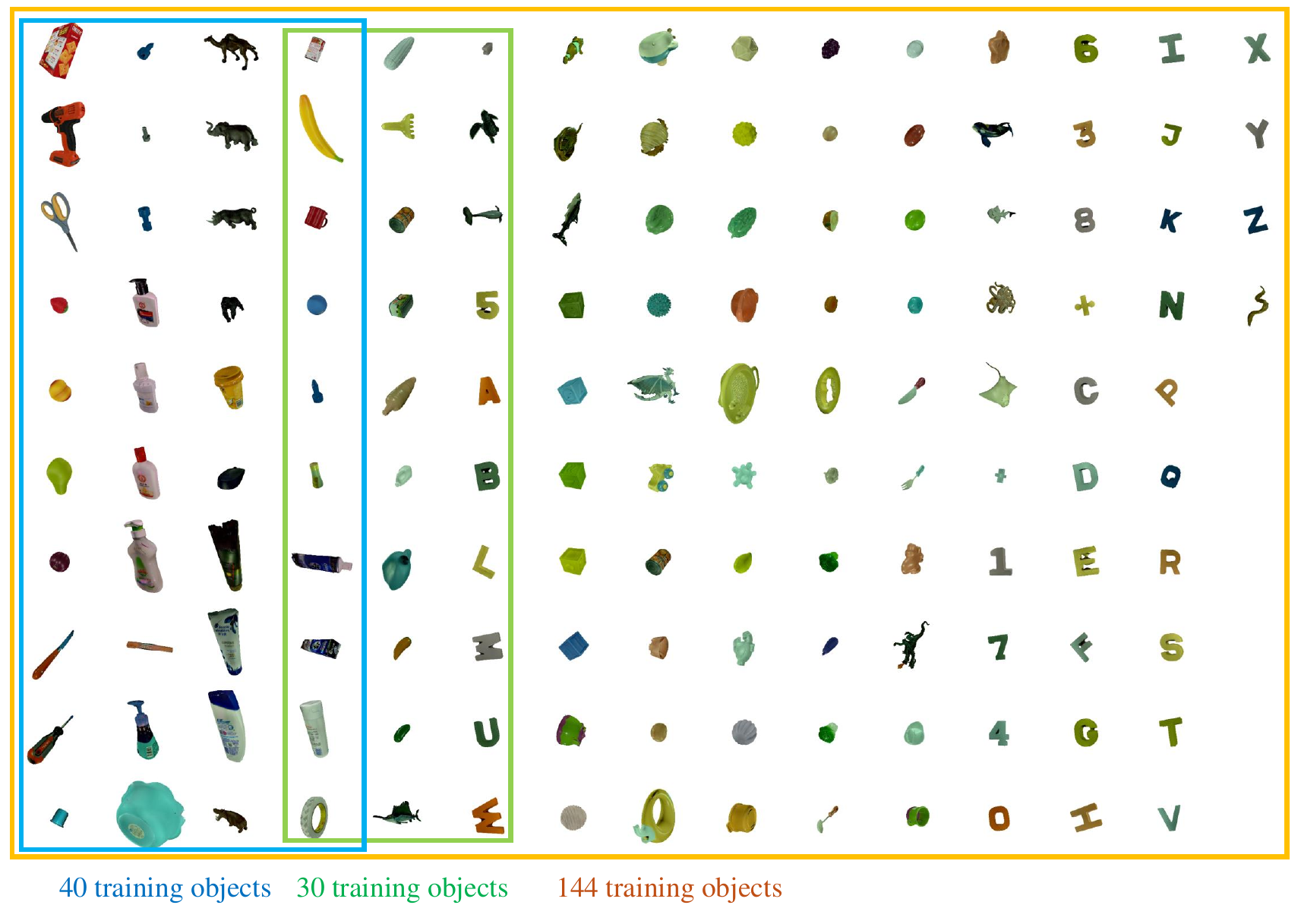}
    \caption{\textbf{Training object set.} The set $\mathbb{L}$ with 144 training objects are enclosed by the orange rectangle, the set $\mathbb{S}$ with 40 training objects are enclosed by the blue rectangle, and the set $\mathbb{T}$ with 30 training objects are enclosed by the green rectangle. Their CAD models are available upon request.
    }
    \label{fig:train_obj}
\end{figure*}

\begin{figure*}
    \centering
    \includegraphics[width=0.8\textwidth]{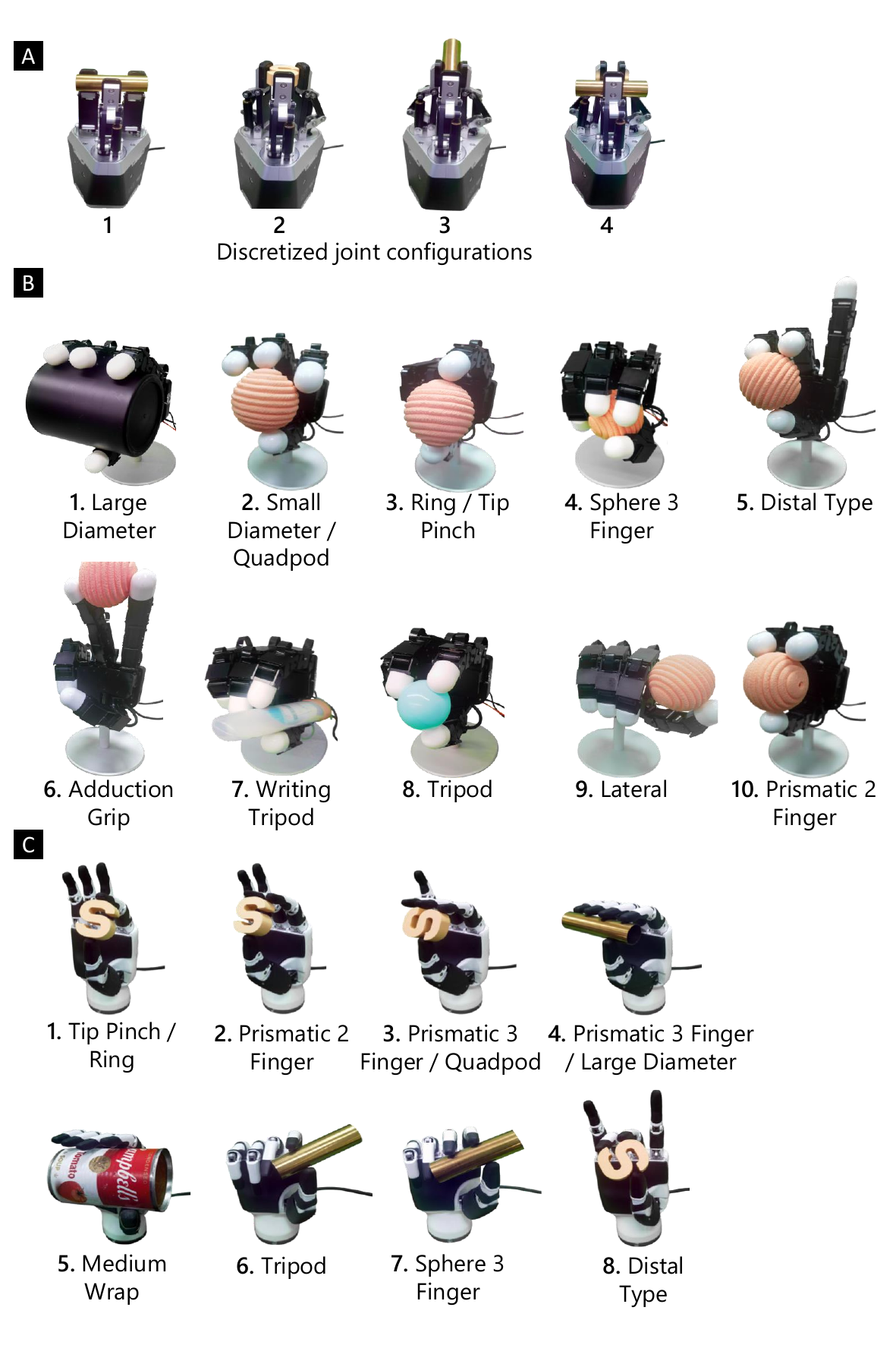}
    \caption{\textbf{Grasp type numbering.} \textbf{(A)}, \textbf{(B)} and \textbf{(C)} give the index numbers of different grasp types for the three-finger, four-finger, and five-finger hands, respectively.
    }
    \label{fig:index}
\end{figure*}

\begin{figure*}
    \centering
    \includegraphics[width=0.8\textwidth]{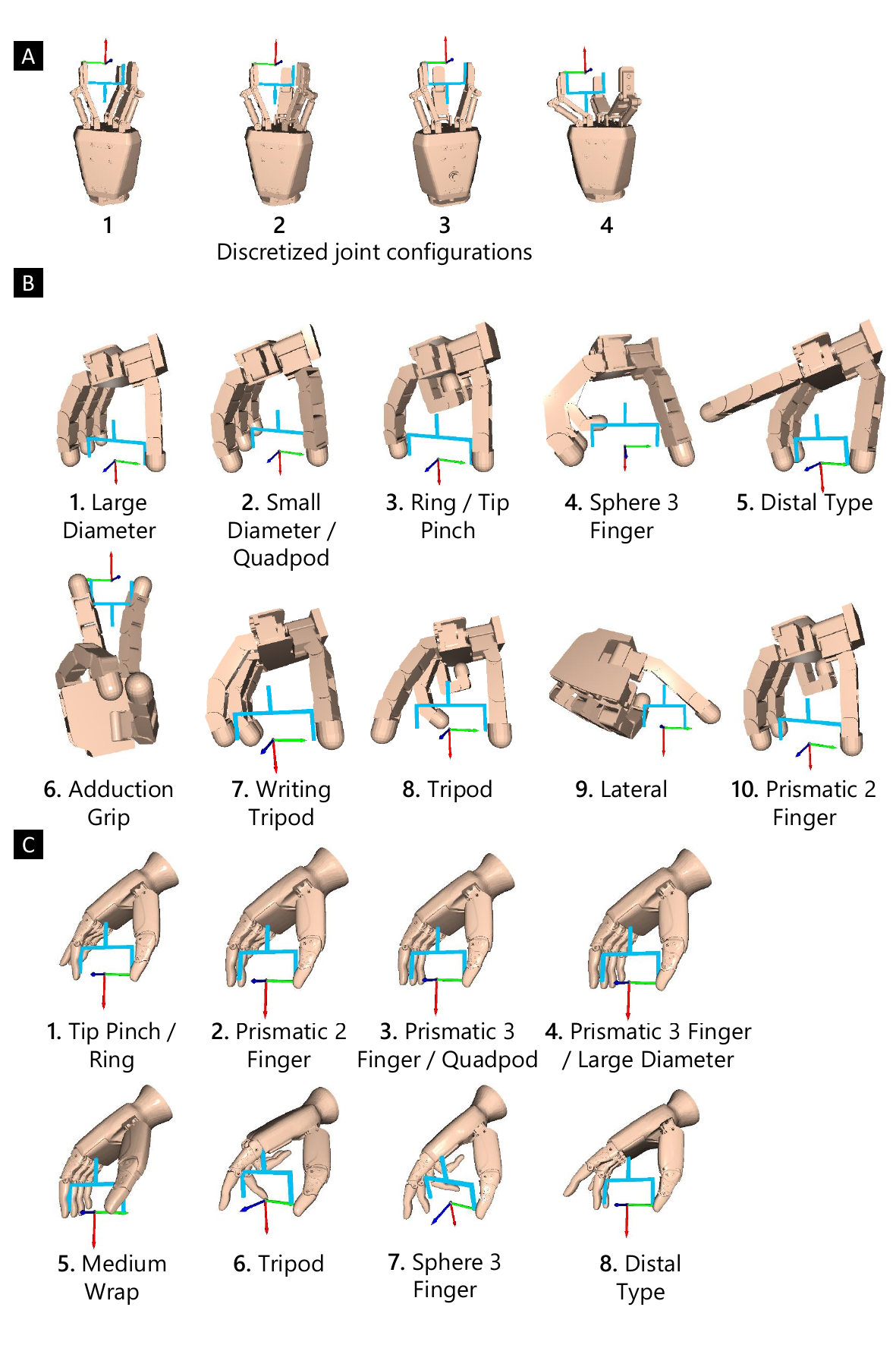}
    \caption{\textbf{Identifying principal closing axis.} We show the designated principal closing axis and corresponding antipodal grasp pose for each predefined grasp type. \textbf{(A)}, \textbf{(B)} and \textbf{(C)} shows the results for the three hands we used.
    }
    \label{fig:principal}
\end{figure*}

\end{document}